\documentclass{article}

\usepackage{microtype}
\usepackage{graphicx}
\usepackage{subcaption}
\usepackage{booktabs} % for professional tables

\usepackage{hyperref}

\usepackage[accepted]{icml2026}

\makeatletter
\renewcommand{\Notice@String}{\textit{Proceedings of the $\mathit{43}^{rd}$ International Conference on Machine Learning}, Seoul, South Korea. PMLR 306, 2026.}
\renewcommand{\printAffiliationsAndNotice}[1]{\global\icml@noticeprintedtrue%
  {\let\thefootnote\relax\footnotetext{\hspace*{-1.8em}\ificmlshowauthors #1\fi%
      \ifdefined\icmlcorrespondingauthor@text
         \\ Correspondence to: \icmlcorrespondingauthor@text.
      \fi
      \\ \Notice@String
    }%
  }%
}
\makeatother

\usepackage{amsmath}
\usepackage{amssymb}
\usepackage{mathtools}
\usepackage{amsthm}

\usepackage{booktabs}
\usepackage{makecell}
\usepackage{graphicx}
\usepackage{pifont}
\usepackage{multirow}
\usepackage{dblfloatfix}
\usepackage{titletoc}
\contentsuse{appendices}{toc}

\newcommand{\xmark}{\ding{55}}
\usepackage[capitalize,noabbrev]{cleveref}

\theoremstyle{plain}

\theoremstyle{definition}

\theoremstyle{remark}

\newcommand{\modelname}{\textsc{Modus}}
\newcommand{\datasetname}{\textsc{Modus-Dataset}}
\newcommand{\method}{\modelname{}} % legacy alias used in some table "Ours" rows

\usepackage{xspace}
\makeatletter
\DeclareRobustCommand\onedot{\futurelet\@let@token\@onedot}
\def\@onedot{\ifx\@let@token.\else.\null\fi\xspace}
\makeatother
\def\eg{\emph{e.g}\onedot}
\def\ie{\emph{i.e}\onedot}

\newcommand{\new}[1]{#1}
\newenvironment{revised}{}{}

\newcommand{\modnew}[1]{#1}

\usepackage[textsize=tiny]{todonotes}

\icmltitlerunning{\modelname{}: Decoder-Only Any-to-Any Modeling of Diverse Modalities}

\begin{document}

\twocolumn[
  \icmltitle{\modelname{}: Decoder-Only Any-to-Any Modeling of Diverse Modalities}



  \icmlsetsymbol{equal}{*}
  \icmlsetsymbol{techadvise}{$\dagger$}

  % Override default affiliation numbering: place Apple at index 2 by
  % pre-creating the @affil* counters in the desired order. Affiliations are
  % otherwise numbered by first appearance among authors (which would put
  % Apple at 4). To revert, delete this \makeatletter block.
  \makeatletter
  \newcounter{@affilepfl}\stepcounter{@affiliationcounter}\setcounter{@affilepfl}{\value{@affiliationcounter}}
  \newcounter{@affilapple}\stepcounter{@affiliationcounter}\setcounter{@affilapple}{\value{@affiliationcounter}}
  \newcounter{@affilku}\stepcounter{@affiliationcounter}\setcounter{@affilku}{\value{@affiliationcounter}}
  \newcounter{@affilcuhk}\stepcounter{@affiliationcounter}\setcounter{@affilcuhk}{\value{@affiliationcounter}}
  \newcounter{@affilunige}\stepcounter{@affiliationcounter}\setcounter{@affilunige}{\value{@affiliationcounter}}
  \newcounter{@affillambda}\stepcounter{@affiliationcounter}\setcounter{@affillambda}{\value{@affiliationcounter}}
  \makeatother

  \begin{icmlauthorlist}
    \icmlauthor{Mingqiao Ye}{equal,epfl}
    \icmlauthor{Zhaochong An}{equal,epfl,ku}
    \icmlauthor{Zhitong Gao}{epfl}
    \icmlauthor{Xian Liu}{cuhk}
    \icmlauthor{Fran\c{c}ois Fleuret}{unige}\\
    \icmlauthor{Chuan Li}{lambda}
    \icmlauthor{Amir Zadeh}{lambda}
    \icmlauthor{Serge Belongie}{ku}
    \icmlauthor{Afshin Dehghan}{apple}
    \icmlauthor{Jesse Allardice}{techadvise,apple}\\
    \icmlauthor{David Mizrahi}{techadvise,apple}
    \icmlauthor{O\u{g}uzhan Fatih Kar}{techadvise,epfl,apple}
    \icmlauthor{Roman Bachmann}{techadvise,epfl,apple}
    \icmlauthor{Amir Zamir}{epfl}
  \end{icmlauthorlist}

  \icmlaffiliation{epfl}{EPFL, Lausanne, Switzerland}
  \icmlaffiliation{apple}{Apple}
  \icmlaffiliation{ku}{University of Copenhagen, Copenhagen, Denmark}
  \icmlaffiliation{cuhk}{The Chinese University of Hong Kong, Hong Kong, China}
  \icmlaffiliation{unige}{University of Geneva, Geneva, Switzerland}
  \icmlaffiliation{lambda}{Lambda AI}

  \icmlcorrespondingauthor{Mingqiao Ye}{mingqiao.ye@epfl.ch}

  \icmlkeywords{Machine Learning, ICML}

  \ificmlshowauthors
  {\centering
  \vspace{0.5em}%
  {\small
  \textsuperscript{1}EPFL\quad
  \textsuperscript{2}Apple\quad
  \textsuperscript{3}University of Copenhagen\quad
  \textsuperscript{4}CUHK\quad
  \textsuperscript{5}University of Geneva\quad
  \textsuperscript{6}Lambda AI\par}%
  \par}
  \fi

  {\centering
  \vspace{0.4em}%
  \url{https://modus-multimodal.epfl.ch/}%
  \par}

    {\centering
    \vspace{0.5em}%
    \captionsetup{type=figure}%
    \includegraphics[width=0.95\linewidth]{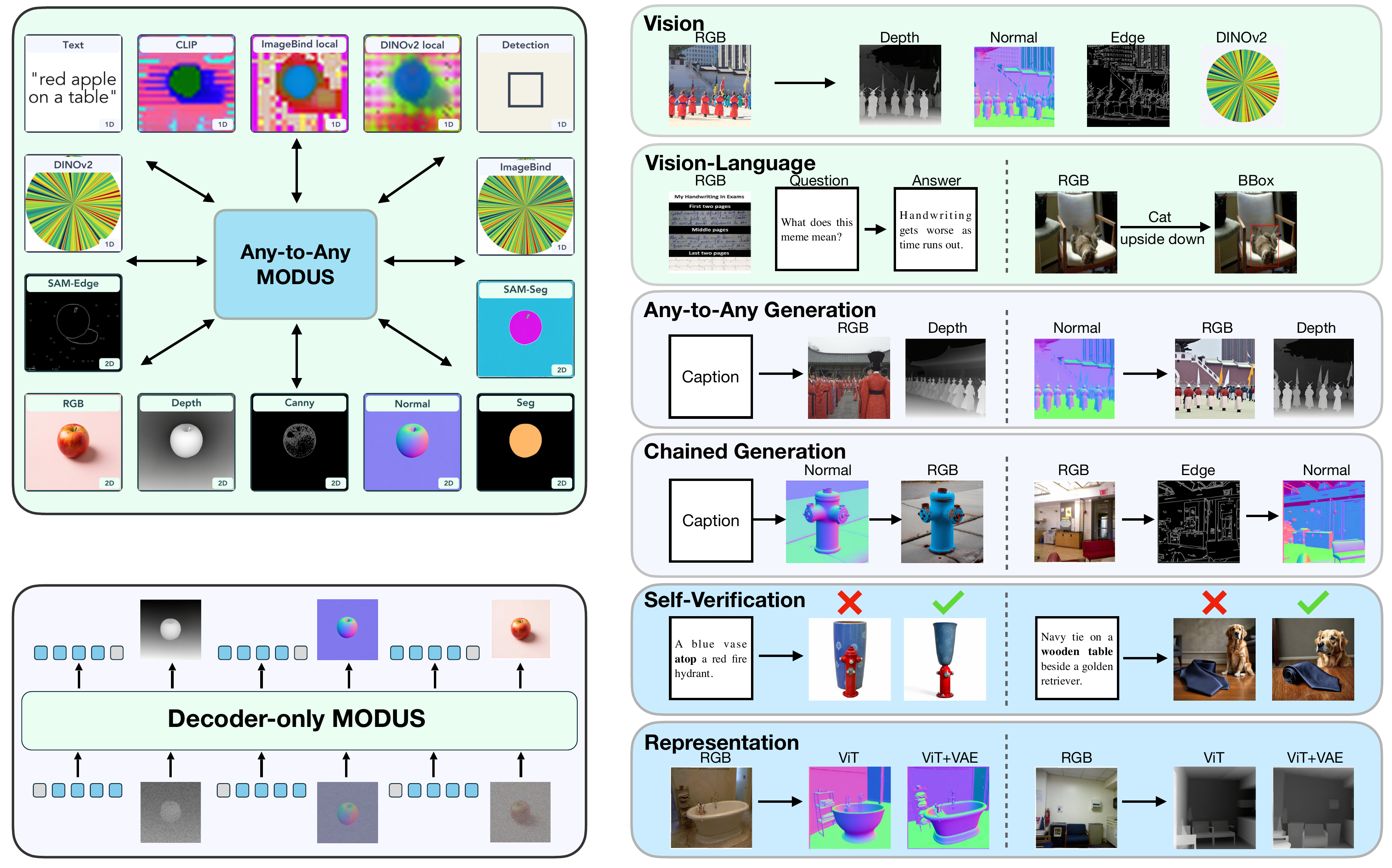}%
    \captionof{figure}{
      (1) Bottom left: \textbf{Decoder-Only.} \modelname{} is a decoder-only model, where all modalities are processed and generated within a single Transformer decoder using a unified autoregressive token sequence without modality-specific heads. (2) Top left: \textbf{Any-to-Any}. \modelname{} performs any-to-any modeling, \ie, predicting any modality given any other, and supports diverse 1D and 2D modalities. 
      (3) Right: \textbf{Multimodal Capabilities.} The resulting framework can support a wide range of capabilities in a single model: core vision tasks such as converting RGB to depth, surface normals, edges, or DINO features; vision–language tasks including VQA, and visual grounding; flexible any-to-any generation across modalities; chained generation through intermediate modalities; self-verification, where one modality can evaluate the output of another; and visual representation composition, where semantic features and reconstruction latents jointly contribute to faithful generations.
    }%
    \label{fig:teaser}\par
  }
  
  \vskip 0.3in
]



\printAffiliationsAndNotice{\icmlEqualContribution\textsuperscript{$\dagger$}Equal technical advising}

% \input{sec/0_abstract}    
% \begin{abstract}
% Decoder-only transformer architectures offer strong scalability, unified training, and zero-shot generalization, but existing models are largely limited to text or image--text modalities. 
% We introduce \modelname{}, an \textbf{any-to-any decoder-only multimodal model} that extends this paradigm to a diverse set of modalities, including geometric, structural, and semantic representations. 
% \modelname{} uses a single decoder to jointly handle 1D sequential modalities and 2D spatial modalities, enabling arbitrary modality-to-modality generation without modality-specific heads or training objectives. 
% As a consequence of this unified design, \modelname{} naturally supports chained generation through intermediate modalities, cross-modal self-verification via consistency checks, and faithful 2D generation by combining semantic and generative representations. 
% Across a wide range of benchmarks, \modelname{} demonstrates strong out-of-the-box performance and flexible multimodal composition within a single model.
% \end{abstract}

\begin{abstract}
%Any-to-any modeling aims to flexibly relate arbitrary modalities within a single model, a requirement that arises across multimodal learning and scientific domains such as ecology and astronomy. 
Any-to-any models predict any modality from any combination of others within a single network -- a formulation used in multimodal vision and vision-language models, and increasingly in scientific domains such as ecology and astronomy.
Existing any-to-any models are typically trained from scratch using encoder-decoder or diffusion architectures, impacting their performance and preventing them from using strong pre-trained decoder-only models as a prior.
In this work, we investigate \textbf{decoder-only any-to-any multimodal modeling}, which treats all modalities symmetrically and supports arbitrary modalities as inputs and outputs without modality-specific heads, losses, or task pipelines.
%\modelname{} performs any-to-any prediction, generating any modality from any combination of others, and supports chained generation through intermediate modalities. We further study cross-modal self-verification, scoring the model's own outputs with another generated modality, and visual representation composition, combining semantic (ViT) and reconstruction (VAE) features for faithful 2D generation.
Because every modality is both an input and an output of the same model, the resulting model, named \modelname{}, can support a range of applications, such as chained generation through intermediate modalities or cross-modal self-verification by scoring the model's own outputs with another generated modality. \modelname{} demonstrates strong out-of-the-box performance %and flexible multimodal composition within a single model.
and is competitive with specialist and multitask baselines using a single model across various benchmarks. All materials are \href{https://modus-multimodal.epfl.ch/}{open-sourced}.
\end{abstract}

\section{Introduction}
\label{sec:intro}

Any-to-any modeling addresses the need to flexibly map between arbitrary modalities. In multimodal learning, this paradigm has been explored by models such as Unified-IO~\cite{lu2022unified,lu2024unified} and 4M~\cite{mizrahi20234m,bachmann20244m}, which aim to unify diverse tasks and modality combinations within a single framework. Beyond that, similar any-to-any settings naturally arise in scientific domains such as genomics~\cite{nair2025nona}, ecology~\cite{sastry2025prom3e}, and astronomy~\cite{parker2025aion}, where diverse data sources can be related in a compositional and reusable manner. %These applications highlight that any-to-any modeling is not a niche capability, but a general and broadly applicable one.

While existing any-to-any models demonstrate that task-agnostic modeling across modalities is feasible, they are typically trained from scratch using encoder–decoder or diffusion-based architectures. This paradigm requires jointly learning modality structures and cross-modal alignment without being able to leverage the priors available in large pretrained decoder-only models, leading to higher training costs and limited scalability. As a result, despite their conceptual generality, these approaches struggle to match the performance, semantic fidelity, generalization, and efficiency of large-scale pretrained foundation models.

Decoder-only architectures provide a compelling paradigm due to their unified next-token prediction objective, strong zero-shot generalization, and efficient inference enabled by mechanisms such as KV-caching. Their success in large-scale language modeling~\cite{hurst2024gpt,touvron2023llama,team2023gemini} and subsequent extension to image–text generation~\cite{lin2023sphinx,chen2024internvl,bagel2025} demonstrates that a single autoregressive decoder can effectively leverage large-scale pretrained priors when extending from text to image modalities. These properties make decoder-only models an attractive candidate for realizing any-to-any multimodal generation at scale.

Despite their strengths, existing decoder-only multimodal models remain largely confined to limited modalities, most commonly treating text and RGB images as privileged inputs or outputs. While recent systems extend decoder-only architectures to additional modalities~\cite{zhan2024anygpt,peng2023kosmos}, they often rely on modality-specific heads, task-dependent losses, or text-centric pivots, limiting their ability to support arbitrary modality-to-modality generation within a single symmetric framework. As a result, the potential of decoder-only models for any-to-any multimodal generation remains underexplored.

We introduce \modelname{}\footnote{\emph{Modus} is the Latin root of the word \emph{modality}.}, a decoder-only architecture for any-to-any multimodal generation. The model treats all modalities symmetrically and avoids modality-specific components, losses, or task pipelines, allowing arbitrary modality pairs to serve as conditioning inputs or generation targets within a single model. Beyond text and RGB images, \modelname{} supports geometric modalities such as depth~\cite{yang2024depth} and surface normals~\cite{ke2025marigold}, structural modalities such as edge maps, semantic modalities such as segmentation masks and object grounding~\cite{ren2024grounded}\modnew{ and detection~\cite{li2022vitdet,fang2024eva02}}, and representational modalities such as DINOv2 features~\cite{oquab2023dinov2}\modnew{, CLIP~\cite{clip}, and ImageBind~\cite{girdhar2023imagebind}}, thereby extending decoder-only modeling to a broader multimodal setting. \modnew{Rather than committing to a fixed set, \modelname{} treats modalities as an open and extensible collection: each family (\eg, edges, segmentation, or learned features) can host multiple instantiations, and additional modalities can be incorporated without architectural changes. Appendix~\cref{tab_supp:modality_overview} lists the full set of supported modalities.}

\modelname{} is a unified framework that builds upon BAGEL~\cite{bagel2025} and can jointly handle 1D sequential modalities and 2D spatial modalities within a single decoder. Discrete modalities such as text, object grounding~\cite{liu2024grounding}\modnew{, detection~\cite{li2022vitdet,fang2024eva02}}, and self-supervised\modnew{ or cross-modal} feature representations~\cite{oquab2023dinov2}\modnew{ (\eg, DINOv2, CLIP~\cite{clip}, ImageBind~\cite{girdhar2023imagebind}, in both global and spatially-local forms)} are modeled autoregressively using next-token prediction, while continuous spatial modalities including RGB images, depth~\cite{yang2024depth}, surface normals~\cite{ke2025marigold}, segmentation masks~\cite{ren2024grounded}, and edge maps are generated using flow matching in latent space~\cite{lipman2022flow}. During training, we randomly sample subsets of modalities as conditioning inputs and use instruction-following prompts to specify the target modality, enabling flexible any-to-any mappings.

\new{To train \modelname{} at scale, we construct \datasetname{}, a 29M-sample multimodal corpus that aligns geometric (depth, surface normals), structural (canny and SAM-based edges), semantic (segmentation, SAM-based masks, grounding boxes, and detection), and representational (DINOv2, CLIP, and ImageBind features) annotations on a shared image base derived from BLIP-3o dataset~\cite{chen2025blip3}. This alignment enables training across arbitrary (input, target) modality pairs, including transformations rarely covered by existing corpora such as depth $\rightarrow$ canny edge or canny $\rightarrow$ surface normal.}
% These pseudo-labels serve as a practical mechanism for exposing the model to rich cross-modal relationships.
To stabilize joint training across multiple modalities, we adopt uniform timestep sampling for flow matching and a staged training procedure that efficiently extends the model to diverse modalities. Uniform timestep sampling improves instruction adherence by providing balanced supervision across diffusion steps and reducing modality confusion during generation. The staged training procedure progressively expands the set of supported modalities and the number of conditioning modalities per sample. Together, these design choices improve cross-modal alignment, long-context handling, and chained any-to-any generation.

We summarize our main contributions as follows:

\textbf{Decoder-only any-to-any modeling.} We show that any-to-any generation across diverse modalities can be realized in a single, fully decoder-only model, without modality-specific heads, losses, or task pipelines, unlike prior any-to-any systems built on encoder--decoder or diffusion backbones trained from scratch.

\textbf{Extending a pretrained model with a simple recipe.} Rather than training from scratch, we introduce a training recipe, uniform timestep sampling and a staged curriculum, that extends a pretrained decoder-only foundation model to any-to-any generation across many modalities, inheriting its priors and scalability. This yields strong out-of-the-box performance, competitive with task-specific specialists, at a fraction of the compute.

\textbf{Diverse modalities at scale, openly released.} \modelname{} extends a single model to $15$ modalities spanning geometry, structure, semantics, and learned representations, trained on the $29$M-sample \datasetname{}. We openly release the dataset together with two model checkpoints.

The unified design further enables capabilities such as chained generation, cross-modal self-verification, and visual representation composition, which we study in \cref{sec:experiments}.

% First, generated outputs can be directly reused as inputs, enabling \emph{chained generation} across multiple modalities (\eg, RGB$\rightarrow$Canny$\rightarrow$Surface Normal) without additional training or architectural changes. Second, the same model can perform \emph{cross-modal self-verification}, where one generated modality (\eg, grounding or VQA) is used to evaluate candidates produced in another modality. 

% We evaluate \modelname{} on a broad range of benchmarks spanning geometric understanding, semantic perception, visual recognition, image generation, and visual question answering. Without any task-specific finetuning, \modelname{} achieves strong zero-shot performance on NYUv2 depth and surface normal estimation~\cite{silberman2012indoor}, COCO instance segmentation and object grounding~\cite{yu2016modeling,lin2014microsoft}, ImageNet retrieval~\cite{deng2009imagenet}, GenEval text-to-image generation~\cite{ghosh2023geneval}, and MMMU visual question answering~\cite{yue2024mmmu}. Beyond competitive performance on individual tasks, these results demonstrate that a single decoder-only model can flexibly support any-to-any generation, chained multimodal composition, and cross-modal verification within a unified framework.

\section{Related Work}

\subsection{Decoder-Only Multimodal Models}
Decoder-only transformer architectures~\cite{hurst2024gpt,touvron2023llama,team2023gemini} have become a dominant paradigm due to their scalability, unified training, and strong zero-shot generalization. Early multimodal extensions such as LLaVA~\cite{liu2023visual}, MiniGPT-4~\cite{zhu2023minigpt}, InternVL~\cite{chen2024internvl}, and Qwen-VL~\cite{bai2025qwen2} couple visual encoders with large language decoders to enable multimodal understanding~\cite{li2026multiple,li2025human}, but their outputs remain text-only. More recent decoder-only models, including Chameleon~\cite{team2024chameleon}, Show-O~\cite{xie2024showo}, EMU~\cite{wang2024emu3}, Janus~\cite{wu2024janus}, Janus-Pro~\cite{chen2025janus}, Janus-Flow~\cite{ma2025janusflow}, BLIP-3o~\cite{chen2025blip3}, and BAGEL~\cite{bagel2025}, unify text and image generation within a single backbone. However, these models primarily focus on text and RGB images and do not support general any-to-any generation across various modalities.

\subsection{Any-to-Any Models}
Any-to-any models are designed to support flexible mappings between multiple modalities within a shared architecture. Encoder--decoder systems such as Unified-IO~\cite{lu2022unified} and 4M~\cite{mizrahi20234m}, as well as unified diffusion models like OneDiffusion~\cite{le2025one}, demonstrate that task-agnostic modeling across modalities is achievable using modality-agnostic objectives. Similar any-to-any trends also arise in other subjects, with recent models such as AION-1~\cite{parker2025aion} in astronomy, ProM3E~\cite{sastry2025prom3e} in ecology, and Nona~\cite{nair2025nona} in functional genomics. However, these approaches are typically trained from scratch, requiring the joint learning of modality structure and cross-modal alignment without leveraging large-scale pretrained foundation models, which can limit scalability and semantic fidelity.
In parallel, LLM-centric approaches such as NExT-GPT~\cite{wu2024next} and AnyGPT~\cite{zhan2024anygpt} integrate diverse modalities by bridging them through a language model, training primarily on Text-to-Any and Any-to-Text tasks. While effective for semantic alignment, this text-centric paradigm constrains native modality-to-modality interactions, as any mapping between two non-textual modalities is mediated by an intermediate textual description, for example expressing depth$\to$image as depth$\to$text$\to$image, which discards fine spatial detail and leaves the generated image only loosely aligned with the input depth. In contrast, \modelname{} emphasizes architectural and procedural unification within a single decoder-only model, enabling flexible any-to-any generation without predefined tasks or reliance on a textual bridge.
\begin{revised}
Concurrent work, Vision-Banana~\cite{visionbanana}, similarly argues that image generation can serve as a universal interface for diverse vision tasks, reinforcing our view that generative modeling is a strong foundation for unified visual capability. \modelname{} extends this perspective to any-to-any modeling across both 1D and 2D modalities, including text, RGB, depth, surface normal, edges, segmentation, grounding, and DINOv2 features, with explicit support for chained generation and cross-modal self-verification.
\end{revised}

% \subsection{Modality-Specific Expert Models}
% Specialized models continue to advance performance in individual visual domains. For geometry, Depth Anything~\cite{yang2024depth}, Marigold~\cite{ke2024repurposing}, DepthFM~\cite{gui2025depthfm}, and Lotus~\cite{he2024lotus} achieve strong depth estimation, while Omnidata~\cite{kar20223d} and GeoWizard~\cite{fu2024geowizard} address surface normal prediction. For semantic understanding and spatial localization, Grounding DINO~\cite{liu2024grounding}, Grounded-SAM~\cite{ren2024grounded}, and GLaMM~\cite{rasheed2024glamm} provide high-quality grounding and segmentation, and self-supervised learners such as DINOv2~\cite{oquab2023dinov2} supply robust visual representations. Although these expert models achieve state-of-the-art results within their respective domains, they remain modality-specific and isolated. \modelname{} instead aims to unify their functional strengths within a single decoder-only framework capable of any-to-any multimodal generation.

\begin{figure*}[t!]
	\centering
\includegraphics[width=1.0\linewidth]{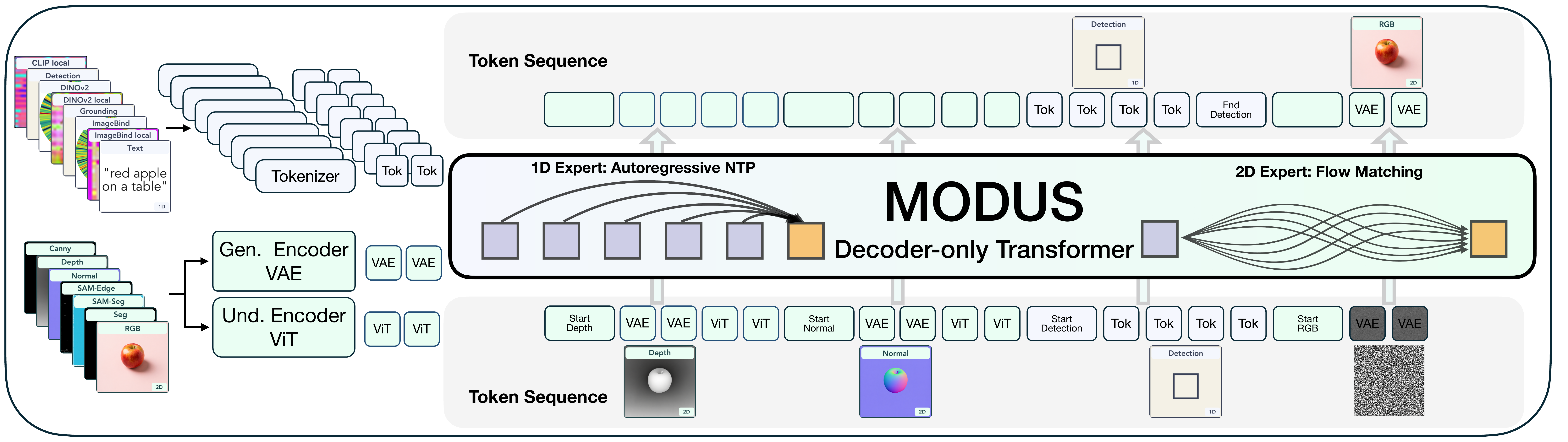}
	% \vspace{-0.2in}
    \caption{\textbf{Overview of the \modelname{} framework.}
    \textbf{(a) Modalities.} \modelname{} supports a diverse and extensible set of modalities. 1D sequential modalities include text, visual grounding boxes, \modnew{object detection,} and \modnew{feature tokens such as} DINOv2\modnew{, CLIP, and ImageBind, in both global and spatially-local forms}. 2D spatial modalities include RGB images, depth, surface normals, segmentation, and canny edges\modnew{, together with SAM-based segmentation and edges}. \modnew{Each modality family can host multiple instantiations, and new modalities can be added without architectural changes.}
    \textbf{(b) Unified tokenization.} All modalities are mapped into a shared sequence with modality instructions. 1D modalities use modality-specific tokenizers. 2D modalities are represented by VAE reconstruction latents and ViT semantic features, with noise added to VAE latents for denoising.
    \textbf{(c) Decoder-only architecture.} \modelname{} employs two experts within a single decoder, a 1D Expert for discrete modalities trained with autoregressive next-token prediction and a 2D Expert for continuous spatial modalities trained with flow matching in latent space. Inside each modality, 1D modalities use causal attention and 2D modalities use bidirectional attention. Among modalities, \modelname{} maintains a shared causal attention context.
    \textbf{(d) Outputs and objectives.} The 1D Expert predicts the next token with a cross-entropy loss. The 2D Expert predicts the velocity field with a mean squared error loss.
    An interactive overview is available at \href{https://modus-multimodal.epfl.ch/\#method}{modus-multimodal.epfl.ch}.
    }

	\label{fig:model}
	% \vspace{-0.2in}
\end{figure*}

\section{Method}

In this section, we present the design of \modelname{}, a unified decoder-only architecture for any-to-any multimodal generation. We first introduce a unified tokenization scheme for diverse modalities~(\cref{sec:modality_unification}), then describe the architectural extensions that enable processing different modalities within a single decoder~(\cref{sec:architecture}). We next present a stabilized training strategy that mitigates modality mixing during joint optimization~(\cref{subsec:training_strategies}). \new{We then describe \datasetname{}, which enables this training paradigm at scale~(\cref{sec:dataset}).} Finally, we describe the inference-time procedures~(\cref{sec:inference}).

\subsection{Unified Tokenization Scheme}
\label{sec:modality_unification}

We design a unified tokenization scheme that represents various modalities in a single sequential format. 
Common modalities can be divided into two groups: \emph{sequential 1D modalities}, which can be modeled through next-token prediction, and \emph{spatial 2D modalities}, which can be generated using flow-matching denoising.

\textbf{1D Modality Representation.}
1D modalities include text, image captions, detection bounding boxes, and feature representations such as DINOv2~\cite{oquab2023dinov2}, CLIP, and ImageBind.
To represent them, we use different discrete tokenizers to obtain token sequences. 
For text inputs such as VQA questions and captions, we use a standard text tokenizer~\cite{bai2023qwen}. 
For bounding boxes, we normalize the coordinates to the range $[0, 999]$ and represent them in the \texttt{x1 y1 x2 y2} format, using 1000 discrete tokens for each coordinate. 
For representation modalities such as DINOv2 features, we adopt an MLP-based tokenizer following~\cite{mizrahi20234m} with a codebook of size 8192.

\textbf{2D Modality Representation.}
2D modalities include RGB images, monocular depth, surface normals, segmentation maps, and canny edge maps. These modalities share spatial structure and can be represented as 2D, RGB-like images.
To capture both semantic and reconstruction-level information, \modelname{} represents each 2D modality using a combination of semantic features and continuous reconstruction latents. Specifically, a pretrained \new{SigLIP-2~\cite{zhai2025siglip}} ViT encoder extracts high-level semantic features, while a pretrained VAE \new{from FLUX~\cite{batifol2025flux}} encodes the same input into continuous latent representations.
The VAE latents serve as the target space for generation and are modeled using a flow-matching objective, enabling high-fidelity spatial synthesis across modalities. For segmentation, we additionally prepend the mask category as text tokens to the sequence.

\textbf{Unified Sequence Format.}
The unified sequence can be formed from any subset of 1D and 2D modalities. \new{Although all modalities share this unified sequence, the tokenization mechanism differs by type: 1D modalities use modality-specific discrete tokenizers, while 2D modalities use the shared ViT--VAE dual representation described above.}
During training, one modality is designated as the generation target, while \new{between $1$ and $3$ of the remaining modalities (sampled randomly)} serve as conditioning inputs\new{; this exposes the model to both single-condition and multi-condition generation within the same training loop}. We define the training objective as:
\vspace{0.5em}
\begin{equation}
[\text{Cond}_1], [\text{Cond}_2], \ldots \rightarrow [\text{Target}]
\end{equation}

% At inference, the model can generate arbitrary target modalities given any combination of conditions and perform chained generation as:
At inference time, the model can generate arbitrary target modalities given any combination of conditioning inputs, and supports chained generation across modalities:
\begin{equation}
\begin{aligned}
[\text{Cond}_1] &\rightarrow [\text{Target}_1] \\
[\text{Cond}_1], [\text{Target}_1] &\rightarrow [\text{Target}_2]
\end{aligned}
\end{equation}

\subsection{Any-to-Any Decoder-Only Architecture}
\label{sec:architecture}

\modelname{} is a decoder-only architecture for \emph{any-to-any multimodal generation}, where arbitrary combinations of modalities can act as conditioning inputs and prediction targets within a single model.
Instead of introducing modality-specific heads, task pipelines, or separate generators, \modelname{} adopts a decoder-only Mixture-of-Transformers structure and incorporates new modalities by unifying their training objectives and token representations within the same sequence modeling framework.
This design preserves the simplicity and strong pretrained priors of decoder-only foundation models, while demonstrating that expressive any-to-any multimodal behavior can be achieved without architectural specialization.

\textbf{Unified Decoder Model.}
Concretely, \modelname{} adapts the pretrained BAGEL-7B~\cite{bagel2025} Mixture-of-Transformers architecture and incorporates two experts within a single autoregressive decoder: a \emph{1D Expert} for sequential modalities and a \emph{2D Expert} for spatial modalities.
The two experts maintain independent parameters but operate over a shared autoregressive token sequence and attend to the same causal self-attention context, such that tokens generated by either expert can condition subsequent tokens.
This unified decoding behavior enables flexible modality composition, forming the architectural basis for any-to-any multimodal generation.

\textbf{Sequential and Spatial Modality Modeling.}
The 1D Expert handles discrete sequential modalities, including text, object grounding, and self-supervised features~\cite{oquab2023dinov2}, using next-token prediction with causal attention.
The 1D Expert captures long-range dependencies and supports semantic reasoning. Given a token sequence $x_{1:N}$, the training objective is
\begin{equation}
\mathcal{L}_{\text{NTP}} = - \sum_{i=1}^{N} \log p_\theta(x_i \mid x_{<i}),
\end{equation}
The 2D Expert models continuous spatial modalities, including RGB images, depth maps, surface normals, segmentation masks, and edge maps, using flow matching~\cite{lipman2022flow}.
Let $x_0$ denote a clean latent representation of a 2D modality encoded by the VAE. A noisy latent $x_t$ is constructed at timestep $t$ by perturbing $x_0$ with Gaussian noise. The model learns a velocity field $v_\theta(x_t, t)$ that matches the time derivative $\dot{x}_t$ of the latent trajectory.
The flow-matching objective is given by
\begin{equation}
\mathcal{L}_{\text{FM}} =
\mathbb{E}_{t, x_0, \epsilon}
\left[
\left\|
v_\theta(x_t, t) - \dot{x}_t
\right\|_2^2
\right].
\end{equation}

\begin{revised}
\textbf{Token Routing.} Each token's path depends on its modality type and role. 1D tokens (text, grounding, DINOv2) flow through the 1D Expert with causal attention and the NTP loss. For 2D modalities, ViT semantic tokens flow through the 1D Expert with intra-modality bidirectional attention, while VAE reconstruction tokens flow through the 2D Expert: clean as a condition, noised at a sampled timestep and supervised by flow matching as the target. Cross-modality attention remains causal so any token conditions all subsequent ones. At inference, condition tokens are encoded first to populate the KV cache before target tokens are decoded.
\end{revised}

\subsection{Training Strategies}
\label{subsec:training_strategies}

\begin{figure}[t!]
  \centering
  \includegraphics[width=\columnwidth]{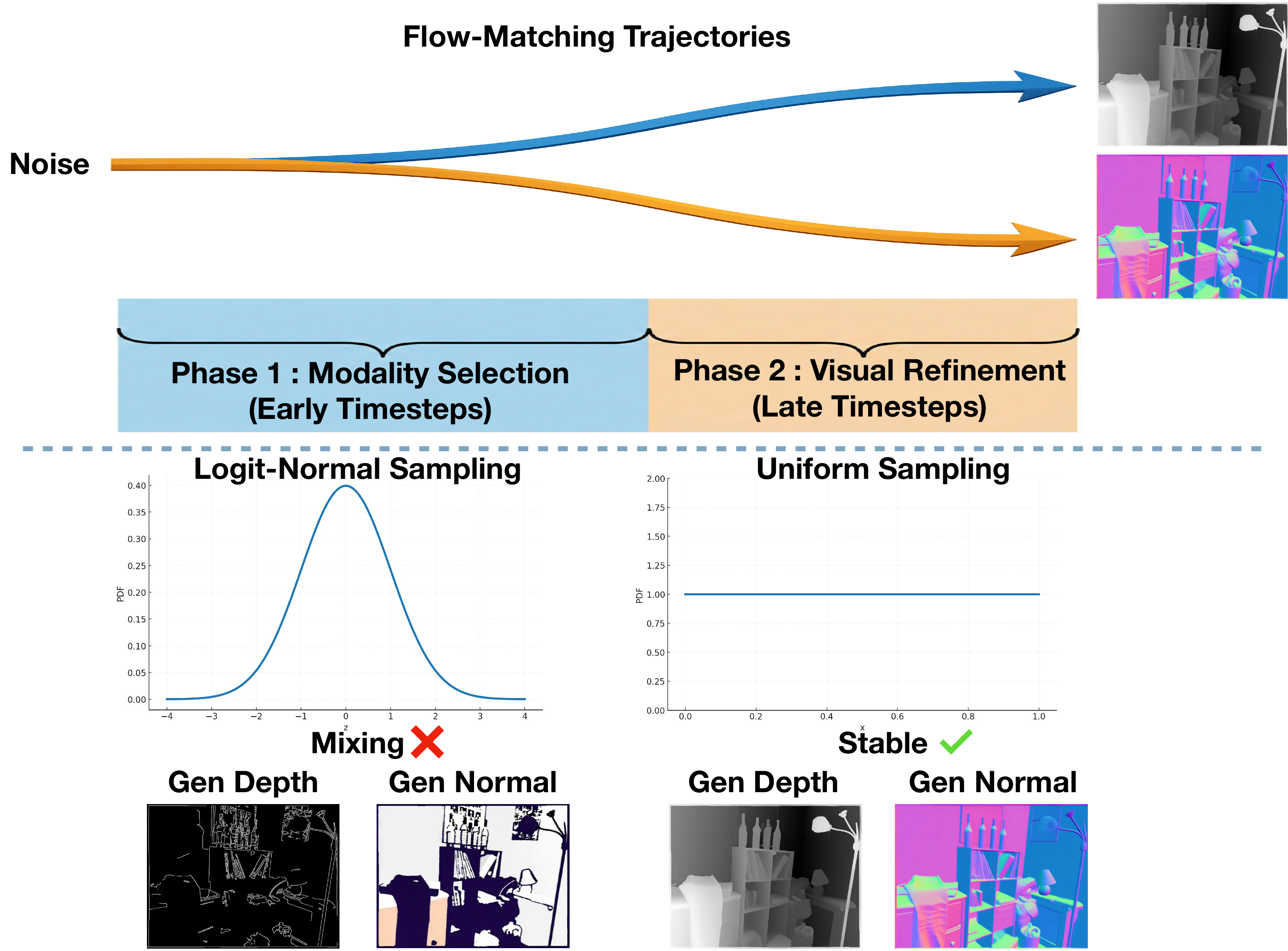}
  \vspace{-0.2in}
  \caption{Top: \textbf{Multimodal Flow Matching.} Early timesteps play a crucial role in determining the target modality, whereas later timesteps primarily refine visual quality. 
  Bottom: \textbf{Timestep Sampling}. Logit-normal sampling undersamples these early steps and oversamples intermediate ones, which causes the model to drift between modalities and produce mixed outputs. Uniform timestep sampling ensures balanced exposure across the full trajectory, enabling the model to reliably follow the target modality instruction and avoid mixing effects.
  }
  \label{fig:modality_mix}
  \vspace{-0.2in}
\end{figure}

\paragraph{Timestep Sampling.}
For flow matching training, we sample a timestep $t$ at each iteration. 
In standard diffusion generation models~\cite{liu2025one,han2025infgen,an2025onestory,an2026vggrpo}, logit-normal sampling is commonly used since it improves generation quality~\cite{Esser2024ScalingRF}. 
However, as shown in~\cref{fig:modality_mix}, in our multimodal decoder-only model, we observe that logit-normal sampling frequently causes a \textit{modality confusion issue}, where the model fails to follow the target instruction, \eg, generating a surface normal map when the intended output is depth. 
We find that early timesteps usually determine the target modality distribution, while later timesteps mainly refine the visual quality. 
Since logit-normal sampling tends to oversample intermediate timesteps and underrepresent early ones, it leads to unstable modality output. 
To address it, we adopt uniform timestep sampling, which provides balanced training across all timesteps and effectively reduces modality confusion.

\textbf{Training Stages.}
We initialize the model from the pretrained BAGEL-7B checkpoint, which supports image and text modalities, and train it through three progressive stages.
In the first stage, training focuses on 1D modalities, including image captions, grounding bounding boxes, and DINOv2 global features~\cite{oquab2023dinov2}. For each sample, one modality is randomly selected as the conditioning input and another as the prediction target, encouraging the model to learn cross-modal alignment and shared priors.
In the second stage, we incorporate 2D modalities such as depth, surface normals, segmentation, and canny edge maps. These spatial modalities exhibit faster convergence, benefiting from strong visual priors.
In the third stage, we increase the number of conditioning modalities per sample to strengthen the model’s ability to handle long-context and multi-condition generation. This stage further improves performance on complex chained generation across diverse modality combinations.

\begin{revised}
\subsection{\datasetname{}}
\label{sec:dataset}
Any-to-any training requires all modalities to be aligned on the same underlying samples; each \datasetname{} training sample is therefore an aligned tuple of one image together with annotations for all its modalities (caption, grounding boxes, detection, depth, surface normals, segmentation, canny edges, and feature-map tokens). We construct \datasetname{} by extending BLIP-3o's image--caption corpus~\cite{chen2025blip3} with aligned annotations from off-the-shelf experts: DepthAnything~\cite{yang2024depth} for depth, Marigold~\cite{ke2025marigold} for surface normals, Grounded-SAM~\cite{ren2024grounded} and SAM~\cite{kirillov2023segment} for segmentation, the Canny operator and SAM~\cite{kirillov2023segment} for edges, GLaMM~\cite{rasheed2024glamm} for grounding boxes, ViTDet~\cite{li2022vitdet} with an EVA-02 backbone~\cite{fang2024eva02} for detection, and DINOv2~\cite{oquab2023dinov2}, CLIP~\cite{clip}, and ImageBind~\cite{girdhar2023imagebind} for representational features. The resulting corpus jointly covers 1D modalities (captions, grounding boxes, detection, and feature-map tokens) and 2D modalities (RGB, depth, surface normals, segmentation, SAM masks, canny edges, and SAM edges), enabling joint training over arbitrary (input, target) modality pairs as well as multi-condition and chained compositions. Detailed dataset statistics are in Appendix~\cref{sec:supp_implementation}.
\end{revised}

\subsection{Inference}
\label{sec:inference}
During inference, the model first encodes all conditioning modalities to construct the KV-cache.
% 1D conditions are processed token by token through the 1D Expert, while 2D conditions are encoded using bidirectional attention over their ViT and VAE features, and the resulting latents are stored in the cache. After the KV-cache is formed, the target instruction and target modality tokens are appended.
The 1D Expert autoregressively predicts discrete tokens, whereas the 2D Expert follows the flow-matching trajectory by predicting velocities over a fixed set of denoising steps. 
Classifier-Free Guidance is applied to improve visual quality for 2D outputs.

Chained generation is enabled by reusing generated outputs as conditioning inputs for subsequent decoding steps. Since all modalities share a unified tokenized representation and are processed by the same model, outputs from one modality can be directly reused to generate another modality, enabling multi-step generation without retraining or architectural changes.
The unified modeling also supports cross-modal self-verification. Given multiple candidate outputs, \modelname{} generates auxiliary modalities conditioned on each candidate, such as grounding or VQA, and selects more coherent results based on cross-modal agreement, without relying on external verifiers or separate scoring models.

% \paragraph{Visual Representation Analysis.}
% Visual representation analysis is enabled by conditioning 2D generation on different combinations of semantic features (\eg, ViT embeddings) and reconstruction latents (\eg, VAE features). This allows controlled comparisons that reveal the complementary roles of semantic alignment and spatial fidelity, with the combined representation yielding more faithful 2D generation.

\section{Experiments}
\label{sec:experiments}

\subsection{Implementation Details}
\new{We train \modelname{} on \datasetname{}~(\cref{sec:dataset}), whose modality alignment supports cross-modal training and inference patterns difficult to study with conventional datasets, such as depth $\rightarrow$ canny or canny $\rightarrow$ surface normal. Training takes approximately 5{,}664 GH200 GPU-hours across three stages ($35$h, $31$h, and $22.5$h on $64$ GPUs).}
We release two \modelname{} checkpoints supporting all $15$ modalities, \texttt{Modus-15modality-14B-A7B} and \texttt{Modus-15modality-77B-A13B}, together with the full \datasetname{} and code.
We present more training details in Appendix~\cref{sec:supp_implementation}.

\subsection{Zero-Shot Generation}

% \begin{figure}[h]
%   \centering
%   \includegraphics[width=0.8\columnwidth]{figures/instseg_vis.pdf}
%   % \vspace{-0.2in}
%   \caption{
%   \textbf{Visualization of Zero-shot RGB$\leftrightarrow$Seg Generation.} This figure will be merged into the matrix any-to-any.
%   % Given either an RGB image or a segmentation mask as input, \modelname{} can generate the corresponding target modality without task-specific fine-tuning. The top row shows RGB$\rightarrow$Seg predictions, and the bottom row shows Seg$\rightarrow$RGB synthesis. These examples demonstrate that the unified decoder-only model learns bidirectional consistency between appearance and instance-level segmentation, capturing both structural layout and fine visual detail. ()
%   }
%   \label{fig:instseg_vis}
%   \vspace{-0.2in}
% \end{figure}

\begin{figure}[t]
  \centering
  \includegraphics[width=1.0\columnwidth]{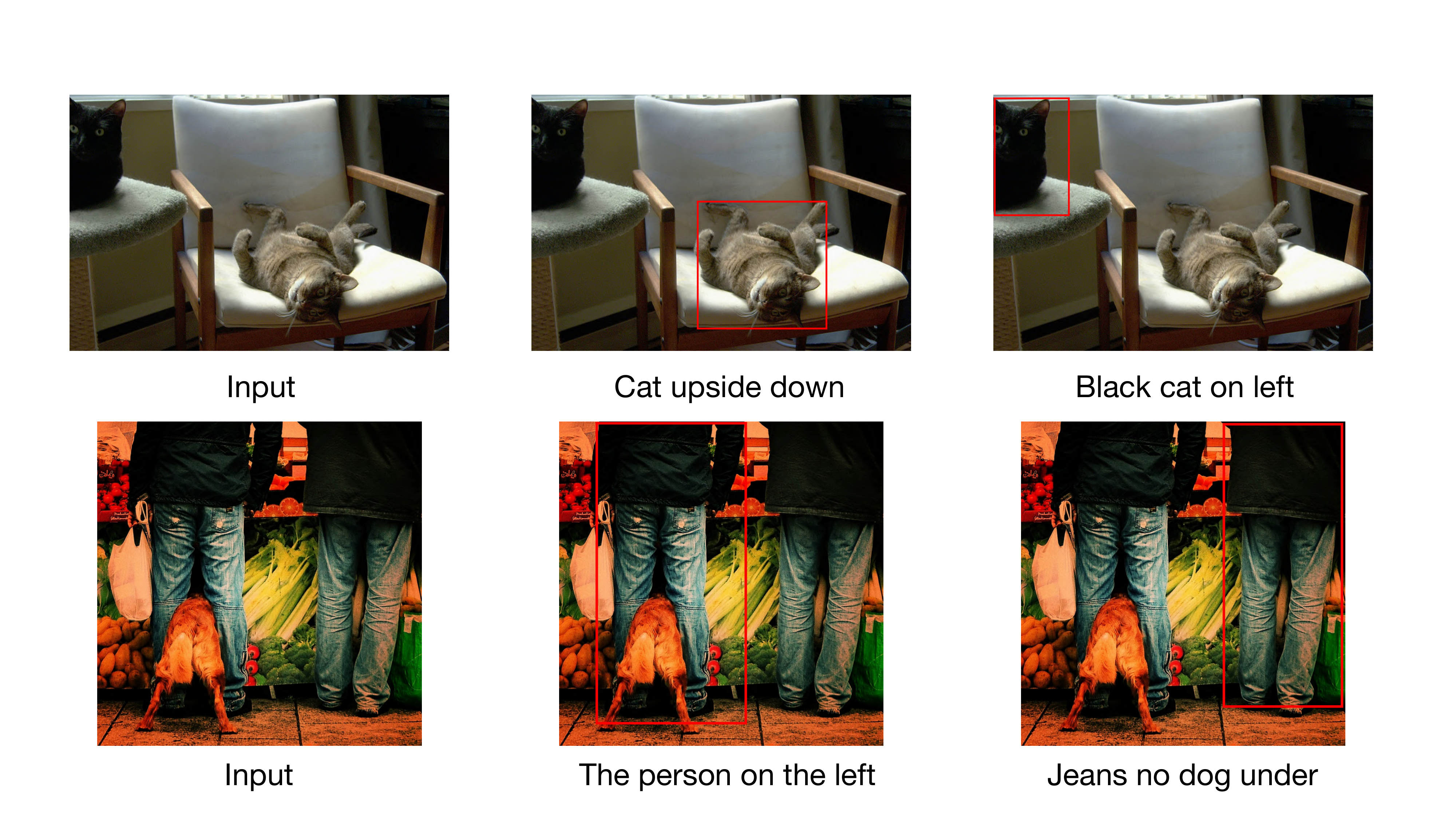}
  \vspace{-0.15in}
  \caption{
    \textbf{Visualization of Zero-shot Grounding Results.} Given an image and a text query, \modelname{} predicts the corresponding region in a zero-shot manner within the unified decoder-only model.
  }
  \label{fig:grounding_vis}
  \vspace{-0.15in}
\end{figure}

\begin{figure}[t]
  \centering
  \includegraphics[width=1.0\columnwidth]{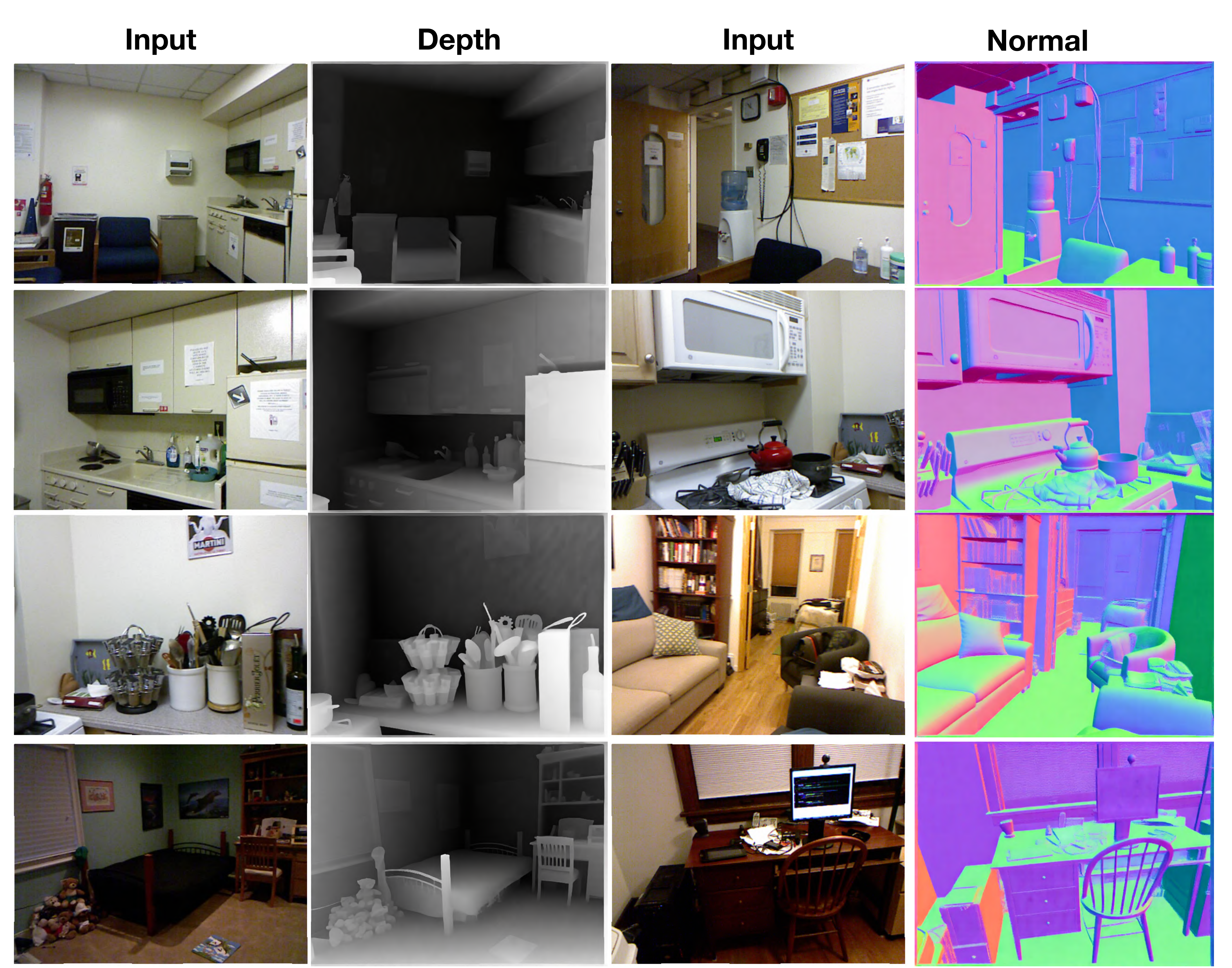}
  \vspace{-0.1in}
  \caption{
  \textbf{Zero-shot Depth and Surface Normal Estimation}. Visualization on the NYUv2 Dataset.
  }
  \label{fig:depth_vis}
  \vspace{-0.2in}
\end{figure}

We qualitatively evaluate the capability of \modelname{} to perform flexible any-to-any multimodal generation across a diverse set of modalities.
Figures~\ref{fig:grounding_vis} and~\ref{fig:depth_vis} present representative zero-shot visualizations demonstrating the model’s unified generative behavior. \modelname{} can take an arbitrary input modality and generate all other target modalities within the same architecture; we visualize this independent any-to-any generation across all modality pairs in Appendix~\cref{subsec:supp_independent} and \cref{fig:any2any_vis}.

\textbf{Visual grounding and geometric understanding.}
Figure~\ref{fig:grounding_vis} shows zero-shot grounding results, where the model localizes text-specified regions in complex natural images. 
Figure~\ref{fig:depth_vis} presents zero-shot depth and surface normal estimation on the NYUv2 dataset. In both cases, \modelname{} generalizes to unseen data and produces geometrically coherent predictions.

\begin{table*}[t!]
\centering
\caption{\textbf{Zero-shot Benchmark across Tasks.} ↑ Higher is better. ↓ Lower is better. \xmark~denotes models that cannot solve the task out-of-the-box. The first group (in gray) reports single-task specialist models, each solving only one task. The next groups present existing any-to-any models, including encoder–decoder and diffusion approaches. The last group shows decoder-only models. \new{\modelname{} extends decoder-only models from image–text settings to diverse modalities and operates on all benchmarks in a zero-shot manner. It remains competitive with multitask any-to-any baselines and decoder-only image--text models, while supporting broader multimodal capabilities (any-to-any, chained generation, cross-modal verification) that prior systems do not.} † results reproduced by us.}
\vspace{-0.2cm}
\setlength{\tabcolsep}{8pt}
\resizebox{\textwidth}{!}{%
\begin{tabular}{llcccccc}
\toprule
\multicolumn{2}{c}{\textbf{Modality}} &
\multicolumn{1}{c}{RGB$\rightarrow$Text} &
\multicolumn{1}{c}{Text$\rightarrow$RGB} &
\multicolumn{1}{c}{RGB$\rightarrow$Depth} &
\multicolumn{1}{c}{RGB$\rightarrow$Surface Normal} &
\multicolumn{1}{c}{RGB$\rightarrow$Grounding} &
\multicolumn{1}{c}{RGB$\rightarrow$DINO} \\
\multicolumn{2}{c}{\textbf{Task}} &
VQA & T2I & Depth Est. & Surface Normal Est. & Grounding & Retrieval \\
\multicolumn{2}{c}{\textbf{Benchmark}} &
\textbf{MMMU}↑ & \textbf{GenEval}↑ & \textbf{DIODE}↓ &
\textbf{NYUv2}↓ & \textbf{RefCOCO$_{val}$}↑ & \textbf{ImageNet}↑ \\
\midrule

% -------- Single-task specialist group (gray) --------
\multirow{6}{*}{\color{gray}Single-task specialist} & \color{gray}DeepSeek-VL2~\cite{wu2024deepseek} & \color{gray}51.1 & \color{gray}\xmark & \color{gray}\xmark & \color{gray}\xmark & \color{gray}\xmark & \color{gray}\xmark \\
& \color{gray}FLUX.1-dev~\cite{batifol2025flux} & \color{gray}\xmark & \color{gray}0.82 & \color{gray}\xmark & \color{gray}\xmark & \color{gray}\xmark & \color{gray}\xmark \\
& \color{gray}DepthAnything2~\cite{yang2024depth} & \color{gray}\xmark & \color{gray}\xmark & \color{gray}0.249 & \color{gray}\xmark & \color{gray}\xmark & \color{gray}\xmark \\
& \color{gray}Marigold~\cite{ke2024repurposing} & \color{gray}\xmark & \color{gray}\xmark & \color{gray}\xmark & \color{gray}16.40 & \color{gray}\xmark & \color{gray}\xmark \\
& \color{gray}GroundingDINO~\cite{liu2024grounding} & \color{gray}\xmark & \color{gray}\xmark & \color{gray}\xmark & \color{gray}\xmark & \color{gray}50.4 & \color{gray}\xmark \\
& \color{gray}DINOv2~\cite{oquab2023dinov2} & \color{gray}\xmark & \color{gray}\xmark & \color{gray}\xmark & \color{gray}\xmark & \color{gray}\xmark & \color{gray}82.1 / 93.9 \\
\midrule

% -------- Multi-task SOTA group --------
\multirow{2}{*}{Encoder-Decoder} & 4M-21~\cite{bachmann20244m}
& \xmark & 0.37 & 0.331 & 37.28 & \xmark & 78.3 / 92.4 \\
& Unified-IO 2~\cite{lu2024unified}
& -- & -- & 0.369 & 28.55 & -- & \xmark \\

\midrule
\multirow{1}{*}{Diffusion} & OneDiffusion~\cite{le2025one}
& \xmark & 0.65 & 0.399 & -- & \xmark & \xmark \\
\midrule

\multirow{5}{*}{Decoder-only} & Bagel$^{\dagger}$~\cite{bagel2025}
& 53.2 & 0.86 & \xmark & \xmark & \xmark & \xmark \\
& Kosmos-2~\cite{peng2023kosmos} & -- & \xmark & \xmark & \xmark & 52.3 & \xmark \\
& Janus-Pro~\cite{chen2025janus} & 41.0 & 0.80 & \xmark & \xmark & \xmark & \xmark \\
& GPT-4o~\cite{hurst2024gpt}
& 69.1 & 0.84 & \xmark & \xmark & \xmark & \xmark \\

% -------- Ours group --------
& \textbf{\modelname{} (Ours)}
& 51.1 & 0.81 & 0.285 & 19.92 & 54.5 & 77.9 / 92.5 \\

\bottomrule
\end{tabular}
}% end resizebox
\vspace{-0.25cm}
\label{tab:multi_task_bench}
\end{table*}

\textbf{System-level comparison.} As shown in~\cref{tab:multi_task_bench}, we present a system-level comparison between \modelname{} and other any-to-any and decoder-only models. 
Encoder–decoder any-to-any architectures are typically trained with masked autoencoding objectives and perform well on reconstruction tasks, 
but they are less effective for open-ended generation. % or text-related tasks. 
Diffusion models excel at image generation but often struggle with text understanding and compositional generation. 
Previous decoder-only models primarily focus on image and text modalities, whereas our \modelname{} unifies all modalities within a single model for any-to-any generation, achieving strong zero-shot performance.
The top group (in gray) lists single-task specialists, each a dedicated model that solves only one task. \modelname{} covers all of these tasks within a single model while remaining competitive with them across depth, surface normals, grounding, and retrieval.
Furthermore, most existing benchmarks are RGB- or text-centric, while \modelname{} supports a broader set of modality transformations beyond these settings, such as transforming canny edge to depth\new{; quantitative results for such non-standard any-to-any and multi-condition mappings (\eg, canny$\to$depth, surface normal$\to$depth, RGB+canny$\to$depth) are reported in \cref{subsec:supp_multi_cond}}.

\subsection{Chained Generation}
\label{sec:chained}

\begin{figure*}[!tb]
  \centering
  \includegraphics[width=\textwidth]{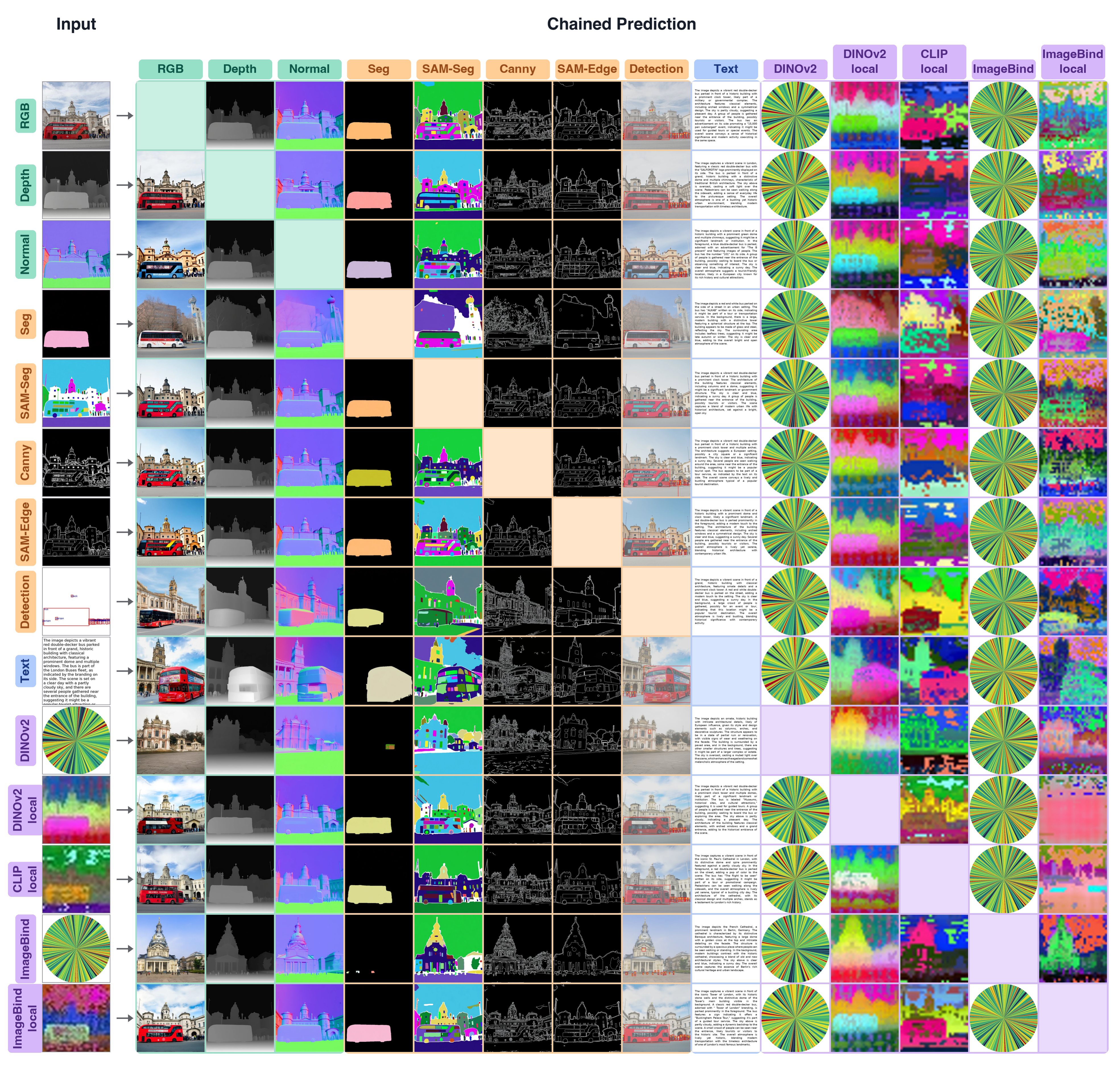}
  \vspace{-0.22in}
  \caption{
  \textbf{Any-to-Any Chained Generation.} Rows are the input modality (leftmost column) and columns are the target; the diagonal (input $=$ target) is left blank. Each target is generated by \emph{chaining}: every newly generated modality is appended to the context used for the next. Because each output is conditioned on the previously generated ones, all targets describe the same underlying scene and stay mutually consistent. The independent counterpart, where each target is generated separately and outputs are free to vary, is shown in Appendix~\cref{fig:any2any_vis}. Interactive visualizations are available at \href{https://modus-multimodal.epfl.ch/\#any-to-any}{modus-multimodal.epfl.ch}.
  }
  \label{fig:chain_gen}
  \vspace{-0.1in}
\end{figure*}

\begin{revised}
\modelname{} can solve a task by \emph{chaining} generation. Because all modalities are produced by the same decoder within a unified token sequence, any generated output can be fed back as a conditioning input for the next step, without retraining or architectural changes. A task can thus be routed through a sequence of intermediate modalities (\eg, RGB $\to$ Canny $\to$ Surface Normal), where each intermediate fixes part of the scene, such as its structure or geometry, before the next modality is generated. As shown in~\cref{fig:chain_gen}, applying this across all input--target pairs makes the intermediate and final outputs describe the same underlying scene and stay mutually consistent, unlike independent generation, where each target is produced separately and the outputs are free to vary~(Appendix~\cref{fig:any2any_vis}).

This composability raises a concrete question: \emph{for a given target, does routing through an intermediate modality help, and which intermediate helps most?} We probe this on surface normal estimation by routing through three candidate intermediates: canny edges, depth, and DINOv2 global features~(\cref{tab:chain_depth}). The results reveal a clear pattern. Spatially-aligned complementary cues help: routing through canny edges improves surface normal prediction, since edges provide pixel-aligned low-level geometry that complements surface orientation. Redundant cues do not help: depth offers no gain over the independent prediction, since depth and surface normals encode closely related geometric information. Non-aligned semantic cues do not help either: DINOv2 global features~\cite{oquab2023dinov2} capture high-level semantics but are neither pixel-aligned nor geometrically informative.

The takeaway is not that ``more steps are better,'' but that chaining lets an intermediate modality supply spatially-aligned structure (\eg, edges) that the target benefits from, a study that requires exactly the modality composability our framework provides. Additional visualizations are in Appendix~\cref{subsec:supp_chained_gen}.

\textbf{Caption-to-Any consistency.} Independent caption-to-target generation can produce visually divergent outputs across modalities (\cref{fig:any2any_vis}), since caption conditioning is inherently sparse and leaves many visual details unspecified. Chained generation directly addresses this: by routing through an intermediate modality (\eg, Text $\to$ Edge $\to$ RGB), the intermediate fixes the spatial layout before generating the final output, producing stronger structural consistency across modalities (see Appendix \cref{subsec:supp_chained_gen}).

\textbf{Inference cost.} Chaining runs one generation per intermediate modality, adding decoding steps in exchange for more consistent outputs. \cref{subsec:supp_inference} reports independent-vs.-chained efficiency and shows the added latency stays modest.
\end{revised}

\begin{table}[t!]
\centering
\caption{\textbf{Chained Generation} on NYUv2 surface normals.}
\vspace{-0.05in}
\label{tab:chain_depth}
\setlength{\tabcolsep}{8pt} % slightly tighter spacing
\small % optional, makes it look balanced with caption text
\resizebox{\columnwidth}{!}{%
\begin{tabular}{lcc}
\toprule
Source $\rightarrow$ Target & Intermediate & NYUv2 Surface Normal ↓ \\
\midrule
RGB $\rightarrow$ Surface Normal & – & 20.02 \\
RGB $\rightarrow$ Depth $\rightarrow$ Surface Normal & Geometry & 20.06 \\
RGB $\rightarrow$ DINO $\rightarrow$ Surface Normal & Semantics & 20.71 \\
RGB $\rightarrow$ Canny $\rightarrow$ Surface Normal & Layout & \textbf{19.87} \\
\bottomrule
\end{tabular}
}
\vspace{-0.1in}
\end{table}

\subsection{Cross-Modal Self-Verification}

Because \modelname{} generates every modality, it can score its own image outputs with another modality it produces, a form of best-of-$N$ verification~\cite{brown2024large,snell2024scaling} that needs no external verifier. We evaluate this on text-to-image generation in~\cref{tab:self_verify}: for each prompt we sample 4 images and keep the one with the highest self-predicted grounding confidence (or VQA answer likelihood) for the requested objects. Both signals improve generation, lifting GenEval from $0.81$ to $0.84$. Additional visualizations are presented in Appendix~\cref{subsec:self-verify}.

% \begin{table}[t!]
% \vspace{-0.0cm}
% \centering
% \caption{Text-to-image generation with self-verification. We apply the verifier score to select the best-of-4 output.}
% \vspace{-0.2cm}
% \setlength{\tabcolsep}{5pt}
% \resizebox{0.5\columnwidth}{!}{
% \begin{tabular}{cc}
% \toprule
% Verifier & GenEval ↑ \\
% \midrule
% --                 & 0.81 \\
% Object Grounding   & 0.82 \\
% VQA                & \textbf{0.83} \\
% \bottomrule
% \end{tabular}
% }
% \label{tab:self_verify}
% \end{table}

\begin{table}[t!]
\vspace{-0cm}
\centering
\caption{\textbf{Cross-Modal Verification}. We apply the verifier score to select the best-of-4 output on text-to-image generation.}
% \vspace{-0.1cm}
\setlength{\tabcolsep}{5pt}
\renewcommand{\arraystretch}{0.95}  % <--- reduce row spacing
\resizebox{0.5\columnwidth}{!}{
\begin{tabular}{cc}
\toprule
Verifier & GenEval ↑ \\
\midrule
--                 & 0.81 \\
Object Grounding   & 0.82 \\
VQA + Grounding    & \textbf{0.84} \\
\bottomrule
\end{tabular}
}
% \vspace{-0.2cm}
\label{tab:self_verify}
\end{table}
\begin{table}[t!]
% \vspace{-0.2cm}
\centering
\caption{\textbf{Visual Representation Composition.} Depth and surface normal estimation conditioned on different 2D feature representations. ViT provides high-level semantic features, while VAE provides low-level reconstruction features.}
% \vspace{-0.2cm}
\setlength{\tabcolsep}{5pt}
\renewcommand{\arraystretch}{0.9}  % <--- reduce row spacing
\resizebox{0.75\columnwidth}{!}{
\begin{tabular}{cccc}
\toprule
ViT & VAE & NYUv2 Depth ↓ & NYUv2 Surface Normal ↓\\
\midrule
$\checkmark$        &     &   15.1 &  35.30\\
 &  $\checkmark$   &    6.9 & 19.96 \\
  $\checkmark$  &  $\checkmark$   &   \textbf{6.5}  & \textbf{19.92}   \\
\bottomrule
\end{tabular}
}
\label{tab:representation}
\end{table}

\begin{figure*}[t]
  \centering
  % \vspace{-0.1in}
  \includegraphics[width=0.6\textwidth]{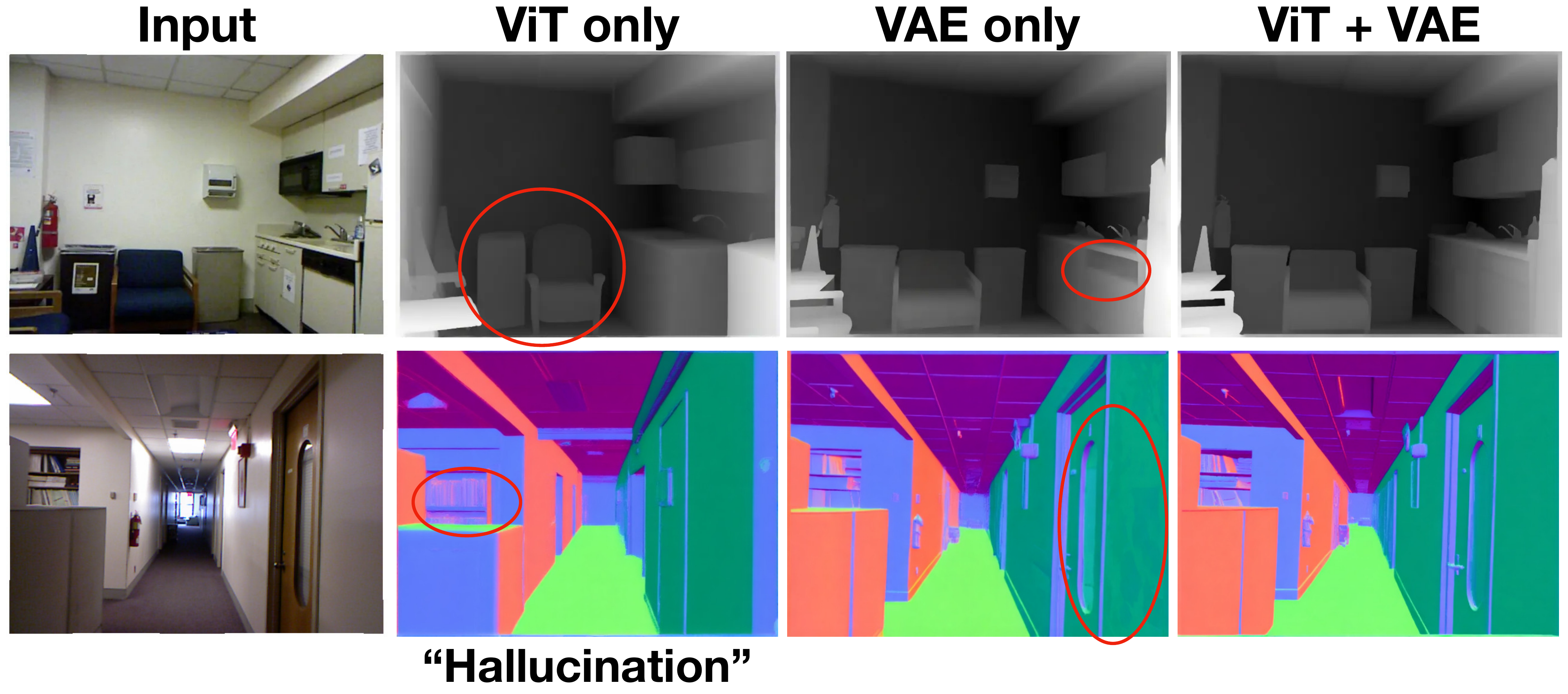}
  \vspace{-0.1in}
  \caption{
  \textbf{Visual Representation Composition.} ViT-only conditioning produces plausible but less faithful outputs with structural changes. VAE features supply low-level details important for geometric consistency. Conditioning on both ViT and VAE yields the most accurate results.
  }
  \label{fig:representation}
  \vspace{-0.1in}
\end{figure*}

\subsection{Visual Representation Composition}
% Our 2D modalities are represented by two complementary features: ViT features capturing high-level semantics and VAE features encoding low-level reconstruction details. During training, generation is conditioned on both features, with dropout applied for classifier-free guidance.

% To analyze the contribution of each feature type, we evaluate the model's performance under three conditioning schemes in~\cref{tab:representation}: ViT-only, VAE-only, and ViT-VAE. The latter performs the best and is our choice. As shown in~\cref{fig:representation}, conditioning solely on ViT features produces plausible target modalities but suffer from hallucinations and misalignments with the input. For instance, generated depth and surface normal maps preserve the rough semantic structure of the input image but may completely alter the detailed geometry of objects. Intriguingly, this characteristic error pattern has also been observed in recent work that evaluated GPT-4o~\cite{ramachandran2025doesgpt4ounderstandvision}, suggesting that GPT-4o may have employed a similar higher-level feature conditioning like ViTs.  Additional visualizations are shown in Appendix~\cref{subsec:supp_visual_representation}.

Our 2D modalities are represented using two complementary features: ViT features that capture high-level semantic structure and VAE features that encode low-level reconstruction details. During training, generation is conditioned on both features, with feature dropout applied to enable classifier-free guidance.
To analyze each representation, we evaluate three conditioning schemes in~\cref{tab:representation}: ViT-only, VAE-only, and ViT–VAE. The combined ViT–VAE setting achieves the best performance. As shown in~\cref{fig:representation}, conditioning solely on ViT features preserves coarse semantic identity but distorts fine-grained geometry; 
for example, in the highlighted regions, the predicted depth still corresponds to a chair, but its geometric shape is altered. 
In contrast, conditioning solely on VAE features maintains local geometric consistency but lacks semantic robustness; 
in the highlighted regions, the black display panel is incorrectly predicted as an empty cavity in the depth map. 

Intriguingly, the same failure mode was reported for GPT-4o~\cite{ramachandran2025doesgpt4ounderstandvision}, hinting that a similar reliance on higher-level features may be at play, though we cannot inspect its internals. Additional visualizations are provided in Appendix~\cref{subsec:supp_visual_representation}.

\subsection{Ablations}

\modnew{Unless otherwise noted, the ablations in this section are conducted under the eight-modality setting (text, RGB, depth, surface normals, segmentation, canny edges, visual grounding, and DINOv2 features); the additional modalities supported by the released model follow the same recipe.}

\textbf{Timestep sampling.}
As discussed in~\cref{subsec:training_strategies} and illustrated in~\cref{fig:modality_mix}, timestep sampling is critical for training 2D experts. In~\cref{tab:ablation_timestep}, we compare logit-normal, mode, and uniform sampling under identical 6B token training budgets.
Logit-normal sampling causes strong modality mixing: 
image-to-depth often collapses into other modalities (\eg, surface normals, edges, RGB, segmentation). 
This arises because all modalities share the same noisy source distribution, and early timesteps contain weak modal boundaries; the model must learn these distinctions in the high-noise regime.
Mode and uniform sampling allocate more probability to early timesteps, giving the model denser supervision where modality boundaries are most ambiguous. As a result, both strategies substantially reduce mixing and yield stronger performance.
Additional visualizations of modality mixing are provided in Appendix~\cref{subsec:supp_modality_mixing}.

\begin{table}[t!]
\centering
\vspace{-0.2cm}
\caption{\textbf{Ablation on 2D Expert Timestep Sampling}. Logit-normal sampling undersamples early timesteps and leads to modality mixing, while uniform sampling provides balanced coverage and yields stable modality-correct outputs.}
\vspace{-0.1cm}
\setlength{\tabcolsep}{6pt} % tighter spacing between columns
\resizebox{0.95\columnwidth}{!}{%
\begin{tabular}{lccc}
\toprule
TimeStep Sampling & GenEval ↑  & NYUv2 Depth ↓ & NYUv2 Surface Normal ↓ \\
\midrule
Logit-Normal & 0.77 & 21.9 & 50.54 \\
Mode         & 0.80 & 8.6 & 22.47 \\
Uniform      & 0.81 & 8.4 & 21.10 \\
\bottomrule
\end{tabular}
}
\vspace{-0.2cm}
\label{tab:ablation_timestep}
\end{table}

\begin{table}[t]
\centering
% \caption{Ablation on training stages.}
\caption{\textbf{Ablation on Training Stages.}
Our three-stage curriculum progressively expands \modelname{}’s capabilities: 
Stage~1 trains 1D modalities, which converge slowly due to lacking strong priors; 
Stage~2 adds additional 2D modalities, improving performance across vision tasks; 
and Stage~3 increases conditioning inputs, enabling multi-condition and chained generation. }
\vspace{-0.05in}
\setlength{\tabcolsep}{5pt}
\renewcommand{\arraystretch}{0.9}
\resizebox{\columnwidth}{!}{
\begin{tabular}{lc|ccccc}
\toprule
Model & \makecell{Training\\Tokens} &
\makecell{\textbf{MMMU}\\↑} &
\makecell{\textbf{GenEval}\\↑} &
\makecell{\textbf{NYUv2}\\↓} &
\makecell{\textbf{Retrieval}\\Top-1 / Top-5 ↑} &
\makecell{Chained\\Gen.} \\
\midrule

Bagel & $\sim$5T
& 53.2 & 0.86 & \xmark & \xmark & \xmark \\

\midrule

\textbf{\modelname{}-Stage1} & 30B
& 51.4 & 0.81 & \xmark & 78.8 / 92.8 & \xmark \\

\textbf{\modelname{}-Stage2} & 20B
& 51.1 & 0.81 & 6.5   & 77.9 / 92.5 & \xmark \\

\textbf{\modelname{}-Stage3} & 15B
& 50.3 & 0.81 & 6.6   & 77.2 / 92.2 & \checkmark \\

\bottomrule
\end{tabular}
}
\vspace{-0.7cm}
\label{tab:ablation_training_stages}
\end{table}

\textbf{Training stages.}
As stated in~\cref{subsec:training_strategies}, we train \modelname{} using a three-stage curriculum. Stage 1 focuses on 1D modalities, including Grounding and DINOv2~\cite{oquab2023dinov2}, which lack pre-existing priors in the model and require longer convergence times. 
% 1D modalities converge slower than 2D modalities when initialized from BAGEL, since BAGEL provides strong vision-language priors that benefit 2D tasks. 
In Stage 2, we incorporate additional 2D modalities, expanding the model's capabilities to various 2D tasks as demonstrated in~\cref{tab:ablation_training_stages}. Finally, in Stage 3, we increase the number of conditioning inputs, enabling the model to perform multi-condition and chained generation. Training losses are in Appendix~\cref{subsec:supp_train_stage}.

\new{\textbf{Additional analyses (appendix).} BAGEL\,+\,per-head baseline (\cref{subsec:supp_ablation_heads}), image--text capability preservation (\cref{subsec:supp_image_text}), per-category GenEval breakdown (\cref{subsec:supp_geneval}), Janus-Flow generalization (\cref{subsec:supp_janus_flow}), and a contamination check (\cref{subsec:supp_contamination}).}

% \subsection{System-level comparison}

% \input{tables/system_level_comparison_new}

% As shown in~\cref{tab:multi_task_bench}, we present a system-level comparison between \modelname{} and other any-to-any and decoder-only models. 
% Encoder–decoder any-to-any architectures are typically trained with masked autoencoding objectives and perform well on reconstruction tasks, 
% but they are less effective for open-ended generation. % or text-related tasks. 
% Diffusion models excel at image generation but often struggle with text understanding and compositional generation. 
% Previous decoder-only models primarily focus on image and text modalities, whereas our \modelname{} unifies all modalities within a single model for any-to-any generation, 
% achieving performance comparable to state-of-the-art models. 
% Furthermore, unlike previous approaches that rely on a single condition for generation, our model supports chained generation with multiple modalities as conditioning inputs.
\section{Conclusion}
\label{sec:conclusion}

We presented \modelname{}, a unified decoder-only model for any-to-any multimodal generation. \modelname{} extends the decoder-only paradigm beyond text and RGB images to support a wide range of modalities within a single model, while avoiding modality-specific heads or task pipelines through unified tokenization.
Stable and scalable multimodal training is achieved using uniform timestep sampling to avoid modality mixing and a staged training procedure that efficiently extends the model to additional modalities and multi-condition settings. 
As a result, \modelname{} supports flexible any-to-any generation, chained generation, cross-modal self-verification, and visual representation composition without additional architectural complexity.
Across diverse benchmarks, \modelname{} demonstrates strong out-of-the-box performance and robust generalization across modality combinations. \new{\modelname{}'s current modality coverage is representative rather than exhaustive, extending to further modalities such as audio or 3D structure is primarily a matter of dataset construction, tokenization, and application demand, which is a promising direction for further work in the community.} This work highlights the potential of decoder-only architectures as a foundation for any-to-any multimodal modeling.

% \clearpage
\section*{Acknowledgments}
We thank Jason Taskov, Kunal Pratap Singh and Muhammad Uzair Khattak for their valuable feedback on earlier versions of the manuscript. We acknowledge Lambda for supporting this paper through academic compute grant program, and a gift from Apple. This work was supported under project ID 43 as part of the Swiss AI Initiative, through a grant from the ETH Domain, with computational resources provided by the Swiss National Supercomputing Centre (CSCS) on the Alps infrastructure. This work was supported by the Swiss AI Initiative (2025 Fellowship Program). This work has also received funding from the Swiss State Secretariat for Education, Research and Innovation (SERI). Zhaochong An and Serge Belongie are supported by funding from the Pioneer Centre for AI, DNRF grant number P1.

\section*{Impact Statement}
This paper aims to advance multimodal representation learning and generation through \modelname{}, a unified any-to-any model. Although the proposed model is not intended for negative use, its general-purpose design and support for diverse modality transformations allow it to be applied beyond the scenarios examined in this work. We position \modelname{} primarily as a research tool for exploring unified multimodal modeling and encourage thoughtful consideration when extending it to broader applications.

% \clearpage
\bibliography{main}
\bibliographystyle{icml2026}

%%%%%%%%%%%%%%%%%%%%%%%%%%%%%%%%%%%%%%%%%%%%%%%%%%%%%%%%%%%%%%%%%%%%%%%%%%%%%%%
%%%%%%%%%%%%%%%%%%%%%%%%%%%%%%%%%%%%%%%%%%%%%%%%%%%%%%%%%%%%%%%%%%%%%%%%%%%%%%%
% APPENDIX
%%%%%%%%%%%%%%%%%%%%%%%%%%%%%%%%%%%%%%%%%%%%%%%%%%%%%%%%%%%%%%%%%%%%%%%%%%%%%%%
%%%%%%%%%%%%%%%%%%%%%%%%%%%%%%%%%%%%%%%%%%%%%%%%%%%%%%%%%%%%%%%%%%%%%%%%%%%%%%%
\newpage
\appendix
\onecolumn

\clearpage
\appendix
\onecolumn

% \let\addcontentsline\oldaddcontentsline

% \section*{\LARGE Appendix}
% \section*{Table of Contents}
% \printapptoc
% \startcontents[appendices]
% \printcontents[appendices]{}{1}{\setcounter{tocdepth}{2}}
% \startcontents[appendices]
% \printcontents[appendices]{l}{1}{\setcounter{tocdepth}{2}}

% --- Manual Appendix Table of Contents (no titletoc needed) ---
\clearpage
\appendix
\onecolumn

\section*{\LARGE Appendix}
\section*{Table of Contents}
\begingroup
\setlength{\parindent}{0pt}
\setlength{\parskip}{3pt}

\textbf{A.} \hyperref[sec:supp_relatedwork]{Additional Related Work}\dotfill\pageref{sec:supp_relatedwork}\\
\textbf{B.} \hyperref[sec:supp_ablation]{Additional Ablations}\dotfill\pageref{sec:supp_ablation}\\
\hspace*{1.5em} \textbf{B.1} \hyperref[subsec:supp_modality_mixing]{Modality Mixing and Timestep Sampling}\dotfill\pageref{subsec:supp_modality_mixing}\\
\hspace*{1.5em} \textbf{B.2} \hyperref[subsec:supp_train_stage]{Training Stages}\dotfill\pageref{subsec:supp_train_stage}\\
\hspace*{1.5em} \textbf{B.3} \hyperref[subsec:supp_train_scratch]{Training from Scratch}\dotfill\pageref{subsec:supp_train_scratch}\\
\hspace*{1.5em} \textbf{B.4} \hyperref[subsec:supp_ablation_heads]{Decoupled vs Unified I/O}\dotfill\pageref{subsec:supp_ablation_heads}\\
\hspace*{1.5em} \textbf{B.5} \hyperref[subsec:supp_janus_flow]{Generalization to a Different Base Model}\dotfill\pageref{subsec:supp_janus_flow}\\

\textbf{C.} \hyperref[sec:supp_evaluation]{More Evaluations}\dotfill\pageref{sec:supp_evaluation}\\
\hspace*{1.5em} \textbf{C.1} \hyperref[subsec:supp_depth]{Depth Estimation}\dotfill\pageref{subsec:supp_depth}\\
\hspace*{1.5em} \textbf{C.2} \hyperref[subsec:supp_grounding]{Referring Object Grounding}\dotfill\pageref{subsec:supp_grounding}\\
\hspace*{1.5em} \textbf{C.3} \hyperref[subsec:supp_image_text]{Image--Text Capability Preservation}\dotfill\pageref{subsec:supp_image_text}\\
\hspace*{1.5em} \textbf{C.4} \hyperref[subsec:supp_multi_cond]{Multi-Condition and Direct Any-to-Any Generation}\dotfill\pageref{subsec:supp_multi_cond}\\
\hspace*{1.5em} \textbf{C.5} \hyperref[subsec:supp_contamination]{Contamination Check}\dotfill\pageref{subsec:supp_contamination}\\
\hspace*{1.5em} \textbf{C.6} \hyperref[subsec:supp_geneval]{GenEval Per-Category Breakdown and Self-Verification}\dotfill\pageref{subsec:supp_geneval}\\
\hspace*{1.5em} \textbf{C.7} \hyperref[subsec:supp_inference]{Inference Efficiency: Independent vs Chained}\dotfill\pageref{subsec:supp_inference}\\

\textbf{D.} \hyperref[sec:supp_visualization]{More Visualizations}\dotfill\pageref{sec:supp_visualization}\\
\hspace*{1.5em} \textbf{D.1} \hyperref[subsec:supp_independent]{Independent Any-to-Any Generation}\dotfill\pageref{subsec:supp_independent}\\
\hspace*{1.5em} \textbf{D.2} \hyperref[subsec:supp_chained_gen]{Chained Generation}\dotfill\pageref{subsec:supp_chained_gen}\\
\hspace*{1.5em} \textbf{D.3} \hyperref[subsec:self-verify]{Self-Verification}\dotfill\pageref{subsec:self-verify}\\
\hspace*{1.5em} \textbf{D.4} \hyperref[subsec:supp_visual_representation]{Visual Representation Composition}\dotfill\pageref{subsec:supp_visual_representation}\\

\textbf{E.} \hyperref[sec:supp_implementation]{Implementation Details}\dotfill\pageref{sec:supp_implementation}\\
\textbf{F.} \hyperref[sec:supp_limitation]{Limitation Discussion}\dotfill\pageref{sec:supp_limitation}\\
\textbf{G.} \hyperref[sec:pseudo_label_vis]{\datasetname{} Examples}\dotfill\pageref{sec:pseudo_label_vis}\\

\endgroup
\clearpage
% --- End manual ToC ---

\newpage

\section{Additional Related Work}
\label{sec:supp_relatedwork}

\textbf{Decoder-only architectures} originated in language modeling and were later extended to vision–language tasks by connecting pretrained LLMs with visual encoders. Early systems such as LLaVA~\cite{liu2023visual}, Qwen-VL~\cite{bai2025qwen2}, and DeepSeek-VL~\cite{wu2024deepseek} follow this pattern: an image encoder (\eg, CLIP~\cite{clip} / SigLIP~\cite{zhai2025siglip}) produces visual features that condition a pretrained GPT-style decoder, enabling image captioning, VQA, and multimodal instruction following. These works established that decoder-only LLMs can serve as a strong multimodal understanding engine, though their outputs remained text-only.

Subsequently, decoder-only transformers were explored for autoregressive image generation. Models such as LlamaGen~\cite{sun2024autoregressive} apply next-token prediction to VQ-based image tokens, while different tokenizers like VAR~\cite{tian2024visual} Infinity~\cite{han2025infinity} and FlexTok~\cite{flextok} increase perceptual fidelity and flexibility. Although these GPT-style generators showed competitive progress, AR image models still face challenges in capturing fine details compared to diffusion-based approaches, motivating further unification of understanding and generation.

A new class of models then aimed to unify image understanding and image generation within a single decoder-only backbone. Early examples include Chameleon~\cite{team2024chameleon} and Show-O~\cite{xie2024showo}, which jointly model text and VQ-tokenized images. Later systems such as EMU-3~\cite{wang2024emu3} and BLIP-3o~\cite{chen2025blip3} improved visual representation quality by integrating pretrained vision encoders or hybrid continuous/discrete tokenizers, enabling stronger semantic reasoning and higher-quality generation.

A major refinement came from Janus~\cite{wu2024janus} and Janus-Pro~\cite{chen2025janus}, which explicitly decouple visual understanding and generation: a semantic encoder (\eg, SigLIP) produces embeddings for recognition tasks, while a VQ-based tokenizer serves the generation branch. JanusFlow~\cite{ma2025janusflow} further augments the generation pathway with a rectified-flow objective, improving realism while keeping the unified decoder-only architecture. Recent large-scale efforts such as BAGEL~\cite{bagel2025} and Hunyuan-Image~\cite{cao2025hunyuanimage} extend this paradigm using Mixture-of-Transformer-Experts, scaling decoder-only multimodal modeling to tens of billions of parameters while supporting both high-quality image synthesis and strong vision–language reasoning.

Decoder-only models are appealing because they scale effectively, inherit strong language understanding, support long context windows, and enable simple parameter sharing across tasks. However, existing models remain largely limited to image–text settings: they may accept other modalities, but they do not support general any-to-any generation across heterogeneous outputs. Our work addresses this gap by extending the decoder-only paradigm beyond photorealistic image and text modalities, supporting additional structured and pixel-dense outputs (\eg, depth, surface normals, segmentation, edges, DINOv2 features, grounding boxes), moving toward a truly unified any-to-any multimodal model.

\paragraph{Modality-Specific Expert Models}
Specialized models continue to advance performance in individual visual domains. For geometry, Depth Anything~\cite{yang2024depth}, Marigold~\cite{ke2024repurposing}, DepthFM~\cite{gui2025depthfm}, and Lotus~\cite{he2024lotus} achieve strong depth estimation, while Omnidata~\cite{kar20223d} and GeoWizard~\cite{fu2024geowizard} address surface normal prediction. For semantic understanding and spatial localization, Grounding DINO~\cite{liu2024grounding}, Grounded-SAM~\cite{ren2024grounded}, and GLaMM~\cite{rasheed2024glamm} provide high-quality grounding and segmentation, and self-supervised learners such as DINOv2~\cite{oquab2023dinov2} supply robust visual representations. Although these expert models achieve state-of-the-art results within their respective domains, they remain modality-specific and isolated. \modelname{} instead aims to unify their functional strengths within a single decoder-only framework capable of any-to-any multimodal generation.

\section{Additional Ablations}
\label{sec:supp_ablation}

\subsection{Modality Mixing and Timestep Sampling}
\label{subsec:supp_modality_mixing}

\begin{figure}[htbp]
  \centering
  \includegraphics[width=0.95\columnwidth]{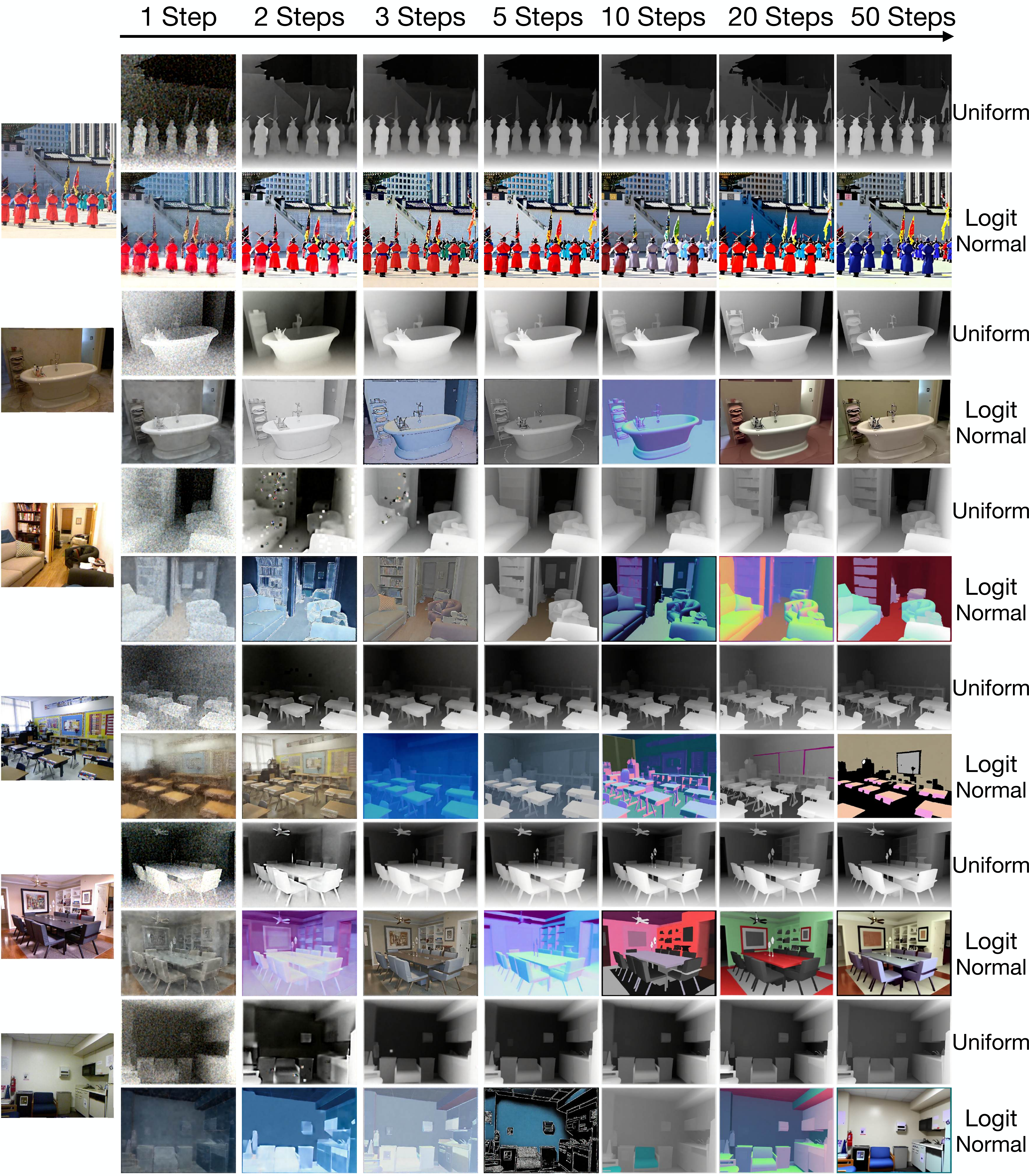}
  \caption{
\textbf{Modality confusion and timestep sampling for depth prediction.}
We compare inference trajectories across timesteps for models trained with uniform and logit-normal sampling. Uniform sampling covers early timesteps well, allowing the model to establish the correct depth modality early and produce stable predictions. Logit-normal sampling undersamples this regime, leading to early-stage modality mixing and failure to recover the correct output. 
  }
  \label{fig_supp:modality_mix_depth}
\end{figure}

\begin{figure}[htbp]
  \centering
  \includegraphics[width=0.95\columnwidth]{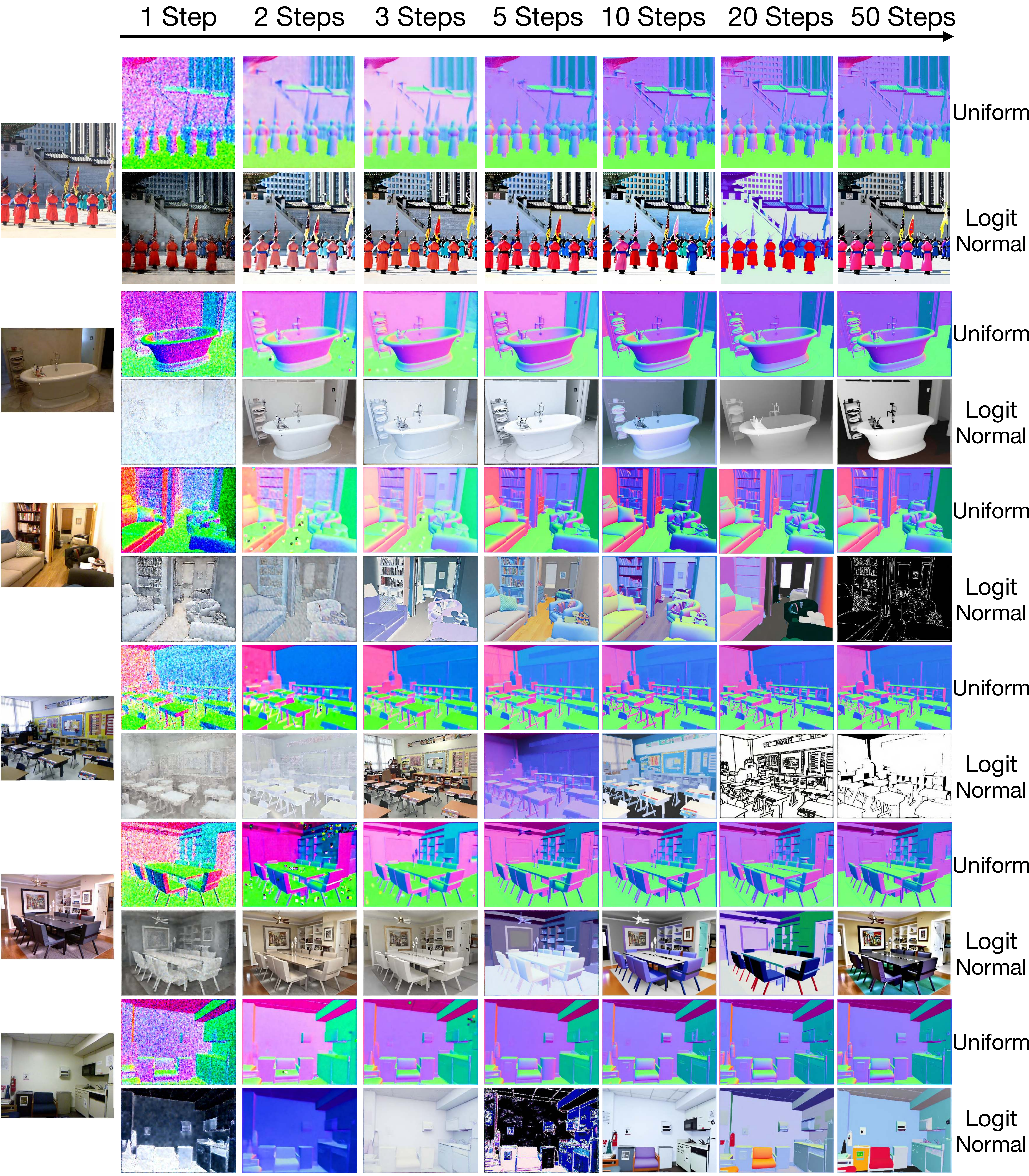}
  \caption{
\textbf{Modality confusion and timestep sampling for surface normal prediction.}
We compare inference trajectories from models trained with uniform and logit-normal sampling. With uniform sampling, the model consistently locks onto the surface normal modality in the early timesteps and refines details thereafter. In contrast, logit-normal sampling provides little supervision in this early region, causing the model to drift across modalities and often fail to produce a clean surface normal map.
  }
  \label{fig_supp:modality_mix_normal}
\end{figure}

We discuss modality mixing and its connection to timestep sampling in Sec.~3.3 of the main paper. With logit-normal sampling, we observe substantial modality confusion, such as cases where a depth-generation prompt results in a surface normal map.  In this section, we provide additional visualizations for a more detailed explanation.
As shown in~\cref{fig_supp:modality_mix_depth} and ~\cref{fig_supp:modality_mix_normal}, we present inference results generated using different numbers of denoising timesteps. We compare two models: one trained with \textbf{uniform} timestep sampling and another trained with \textbf{logit-normal} sampling.

For the model trained with uniform sampling, we observe that the predictions remain stable across timesteps for both depth and surface normal estimation. Even with only a few early timesteps, the model already produces a reasonable target modality, while later timesteps primarily refine structural details. This behavior arises because uniform sampling provides balanced coverage over the entire trajectory during training, ensuring that the model frequently sees the early-timestep regime where modality selection is determined. As a result, the model reliably commits to the correct target modality early in the trajectory and avoids modality confusion during inference.

In contrast, logit-normal sampling places most probability mass on middle (cleaner) timesteps and significantly undersamples the early, noisier region during training. Since the model receives far fewer updates at these early timesteps, it struggles to correctly infer the target modality when the signal is still ambiguous. As shown in our visualizations, this often leads to unstable or mixed-modality outputs at early timesteps, and the model may fail to recover even as the trajectory progresses. This explains why logit-normal sampling, while suitable for single-modality diffusion models, can amplify modality confusion in multi-modality flow prediction.

Overall, these results highlight the importance of sufficient early-timestep coverage for reliable target-modality selection. Uniform sampling provides this property, whereas logit-normal sampling may require additional regularization or curriculum strategies to mitigate modality confusion.

\clearpage

\subsection{Training Stages}
\label{subsec:supp_train_stage}

\begin{figure}[h]
  \centering
  \includegraphics[width=1\columnwidth]{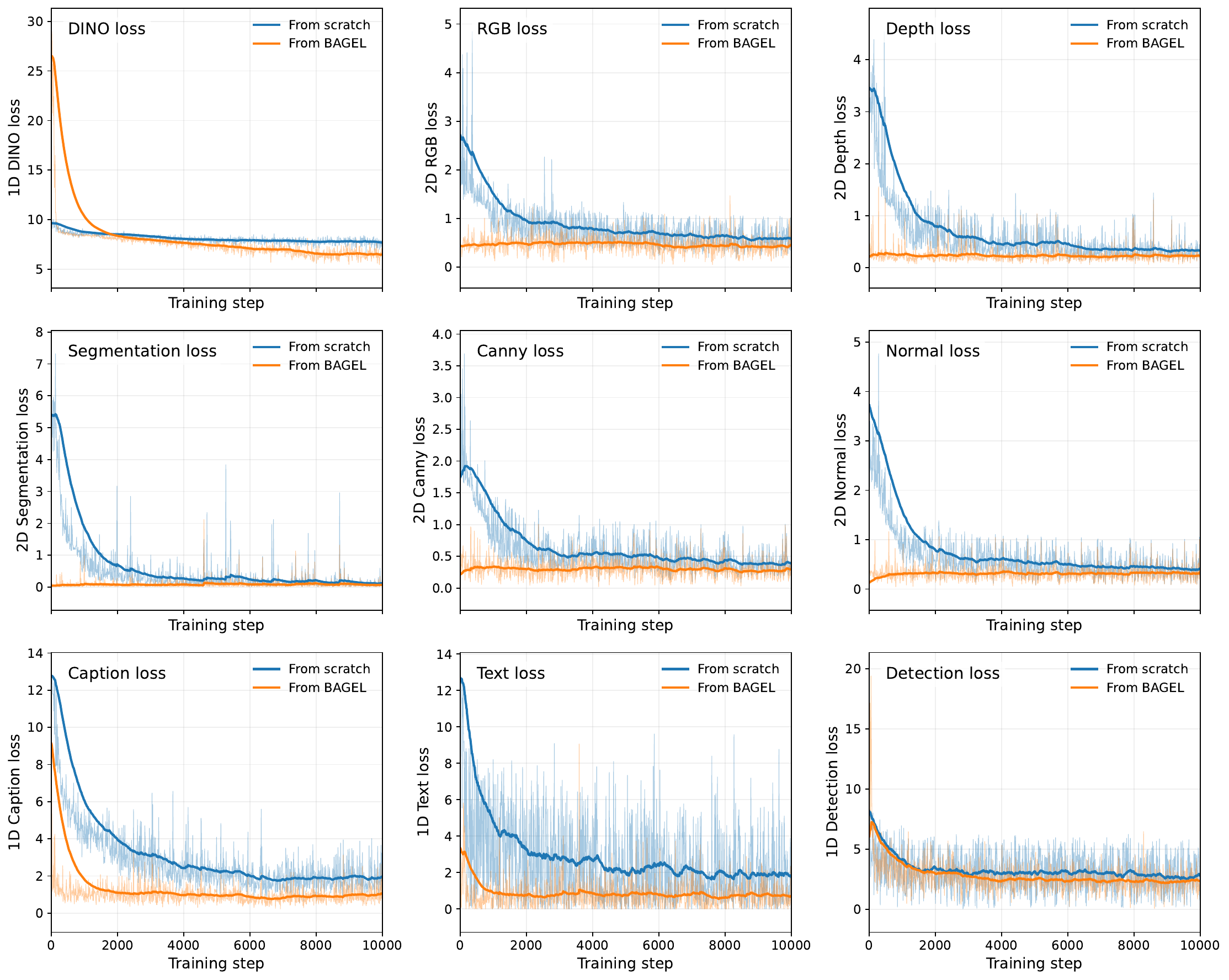}
  \caption{
  \textbf{Training losses across 1D and 2D modalities.} Loss curves for representative modalities during \modelname{} training. All modalities converge stably under the unified training regime, and initializing from BAGEL accelerates convergence across the board.
  }
  \label{fig_supp:train_losses}
\end{figure}

\begin{table}[htbp]
\centering
\caption{Ablation on training stages. We adopt a three-stage training scheme that gradually incorporates 1D and 2D modalities and enables multi-conditioned generation in the final stage.} 
% We also compare with training from scratch, showing that initializing from image–text priors leads to more efficient training
\setlength{\tabcolsep}{8pt}
\resizebox{\columnwidth}{!}{%
\begin{tabular}{lc|cccccccc}
\toprule
 & & 
\multicolumn{1}{c}{RGB$\rightarrow$Text} &
\multicolumn{1}{c}{Text$\rightarrow$RGB} &
\multicolumn{1}{c}{RGB$\rightarrow$Depth} &
\multicolumn{1}{c}{RGB$\rightarrow$Surface Normal} &
\multicolumn{1}{c}{RGB$\rightarrow$Det} &
\multicolumn{1}{c}{RGB$\rightarrow$Seg} &
\multicolumn{1}{c}{RGB$\rightarrow$DINO} \\
 &  Training &
VQA & T2I & Depth Est. & Surface Normal Est. & Grounding & Inst. Segmentation & Retrieval & Chained \\
 & Tokens &
\textbf{MMMU}↑ & \textbf{GenEval}↑ & \textbf{NYUv2}↓ &
\textbf{NYUv2}↓ & \textbf{RefCOCO$_{val}$}↑ & \textbf{COCO}↑ & \textbf{ImageNet}↑ & Generation\\
\midrule

Bagel & $\sim$ 5T
& 53.2 & 0.86 & \xmark & \xmark & \xmark & \xmark & \xmark & \xmark \\
\midrule

% -------- Ours group --------
\textbf{\modelname{}-Stage1} & 30B
& 51.4 & 0.81 & \xmark & \xmark & 56.5 & \xmark & 78.8 / 92.8 & \xmark \\
\textbf{\modelname{}-Stage2} & 20B
& 51.1 & 0.81 & 6.5 & 19.92   & 54.5 & 24.9 & 77.9 / 92.5 & \xmark \\
% \textbf{\modelname{}-Stage3}
% & 50.3 & 0.81 & 6.5 & 20.35 & 58.1 & 23.8 & 77.2 / 92.2 \\
\textbf{\modelname{}-Stage3} & 15B
& 50.3 & 0.81 & 6.6 & 20.02 & 53.8 & 25.0 & 77.2 / 92.2 & $\checkmark $\\
\midrule
\textcolor{gray}{\modelname{}-Scratch} & \textcolor{gray}{50B} & \textcolor{gray}{24.2} & \textcolor{gray}{N/A} & \textcolor{gray}{18.9} & \textcolor{gray}{40.21} & \textcolor{gray}{9.7} & \textcolor{gray}{1.7} & \textcolor{gray}{67.6 / 88.4} & \textcolor{gray}{\xmark} \\
\bottomrule
\end{tabular}
}% end resizebox
% \vspace{-0.2in}
\label{tab_supp:ablation_training_stages}
\end{table}

% \begin{table}[htbp]
% \centering
% \caption{Ablation on training stages.}
% \setlength{\tabcolsep}{10pt}
% \resizebox{1\columnwidth}{!}{%
% \begin{tabular}{lc|ccccc}
% \toprule
%  & Training & 
% VQA & T2I & NYUv2 Depth ↓ & Retrieval (ImageNet) ↑ & Chained Gen. \\
% Model & Tokens &
% \textbf{MMMU}↑ & \textbf{GenEval}↑ & \textbf{NYUv2}↓ &
% \textbf{Top-1 / Top-5}↑ & \\
% \midrule

% Bagel & $\sim$5T
% & 53.2 & 0.86 & \xmark & \xmark & \xmark \\

% \midrule

% \textbf{\modelname{}-Stage1} & 30B
% & 51.4 & 0.81 & \xmark & 78.8 / 92.8 & \xmark \\

% \textbf{\modelname{}-Stage2} & 20B
% & 51.1 & 0.81 & 6.5 & 77.9 / 92.5 & \xmark \\

% \textbf{\modelname{}-Stage3} & 15B
% & 50.3 & 0.81 & 6.6 & 77.2 / 92.2 & $\checkmark$ \\

% \bottomrule
% \end{tabular}
% }
% % \vspace{-0.2in}
% \label{tab:ablation_training_stages}
% \end{table}

% \begin{table}[t]
% \centering
% \caption{Ablation on training stages.}
% \setlength{\tabcolsep}{6pt}
% \renewcommand{\arraystretch}{0.9}
% \resizebox{\columnwidth}{!}{
% \begin{tabular}{lc|ccccc}
% \toprule
% Model & Training Tokens &
% \textbf{MMMU}↑ & \textbf{GenEval}↑ & \textbf{NYUv2}↓ &
% \textbf{Top-1 / Top-5}↑ & Chained Gen. \\
% \midrule

% Bagel & $\sim$5T
% & 53.2 & 0.86 & \xmark & \xmark & \xmark \\

% \midrule

% \textbf{\modelname{}-Stage1} & 30B
% & 51.4 & 0.81 & \xmark & 78.8 / 92.8 & \xmark \\

% \textbf{\modelname{}-Stage2} & 20B
% & 51.1 & 0.81 & 6.5   & 77.9 / 92.5 & \xmark \\

% \textbf{\modelname{}-Stage3} & 15B
% & 50.3 & 0.81 & 6.6   & 77.2 / 92.2 & \checkmark \\

% \bottomrule
% \end{tabular}
% }
% \label{tab:ablation_training_stages}
% \end{table}

As shown in~\cref{tab_supp:ablation_training_stages}, we train \modelname{} using a three-stage curriculum. In the first stage, we introduce only the new 1D modalities so the model can learn modalities without strong priors, such as grounding and DINOv2 feature tokens, over a longer period of iterations. In the second stage, we incorporate the 2D modalities and allow the model to jointly learn the new 1D modalities together with the 2D modalities that come with stronger priors, enabling more efficient training. In the final stage, we enable multi-conditioned generation by mixing arbitrary combinations of modalities, allowing the model to perform any input–output transformation in a unified way. This progressive curriculum stabilizes optimization and helps the model adapt to the full set of modalities step by step. From~\cref{tab_supp:ablation_training_stages}, we observe that the model can be extended to new modalities efficiently while preserving performance on existing ones, and it can even support multi-conditioned chained generation.

\subsection{Training from Scratch}
\label{subsec:supp_train_scratch}

As shown in~\cref{fig_supp:train_losses} and~\cref{tab_supp:ablation_training_stages}, we also train \modelname{} from scratch for both 1D and 2D experts to assess the role of unified MLLM initialization. Training from scratch leads to noticeably slower convergence, especially for text–image related modalities, and results in lower performance on most tasks.

These results indicate that unified MLLM initialization provides useful multimodal priors that improve optimization stability and overall performance in the any-to-any setting. Among all tasks, DINOv2 feature prediction shows the smallest difference between the two training regimes, which is consistent with its lack of semantic priors in both settings.

\begin{revised}
\subsection{Decoupled vs Unified I/O}
\label{subsec:supp_ablation_heads}

Many recent unified-modality models adopt modality-specific output heads to interface different modalities. To isolate the effect of \modelname{}'s unified-sequence design from other factors, we train a controlled baseline that replaces the unified representation with small per-modality prediction heads (a 2D projector and head, a grounding regression head, and a DINO head), while keeping the BAGEL-7B initialization, \datasetname{} training data, and uniform timestep sampling identical.

As shown in \cref{tab_supp:ablation_heads}, the decoupled variant remains close to \modelname{} on MMMU (50.4 vs 51.1) but lags significantly on geometric tasks: 0.322 vs 0.285 on DIODE depth and 50.75 vs 19.92 on NYUv2 surface normal estimation. The large gap on surface normal estimation suggests that per-head decoupling fragments the representation space and weakens cross-modal transfer, whereas the unified sequence preserves cross-modal alignment without requiring per-head output design.
\end{revised}

\begin{table}[t]
\centering
\caption{\textbf{Decoupled vs Unified I/O ablation.} Both variants share BAGEL-7B initialization, \datasetname{} training data, and uniform timestep sampling. The decoupled variant replaces the unified sequence representation with small per-modality prediction heads. The unified-sequence formulation avoids fragmenting the representation across per-modality heads, with the largest gain on surface normal estimation. $\uparrow$ Higher is better. $\downarrow$ Lower is better.}
\label{tab_supp:ablation_heads}
\small
\setlength{\tabcolsep}{6pt}
\begin{tabular}{lccc}
\toprule
Variant & MMMU $\uparrow$ & DIODE $\downarrow$ & NYUv2 Surface Normal $\downarrow$ \\
\midrule
BAGEL + Specific Heads (decoupled) & 50.4 & 0.322 & 50.75 \\
\textbf{\modelname{} (unified sequence)} & \textbf{51.1} & \textbf{0.285} & \textbf{19.92} \\
\bottomrule
\end{tabular}
\end{table}

\begin{revised}
\subsection{Generalization to a Different Base Model}
\label{subsec:supp_janus_flow}

To assess whether the \modelname{} training recipe generalizes beyond BAGEL, we apply the same pipeline (unified-sequence tokenization, uniform timestep sampling, staged training, \datasetname{} data) to Janus-Flow-1.3B~\cite{ma2025janusflow} as a base model. As shown in \cref{tab_supp:janus_flow_generalization}, this extension preserves most of Janus-Flow's MMMU performance ($27.1$ vs $29.3$) while adding depth and surface normal estimation capabilities ($0.304$ DIODE, $36.53$ NYUv2 surface normal). Janus-Flow originally requires $19.2$k GH200 hours of pretraining; extending it with the \modelname{} pipeline takes an additional $1.5$k GH200 hours, roughly $13\times$ less than the original pretraining. This suggests that the unified-sequence and timestep-sampling choices transfer across base architectures rather than being specific to BAGEL.
\end{revised}

\begin{table}[t]
\centering
\caption{\textbf{Generalization of the \modelname{} pipeline to a different base model.} Applying the \modelname{} training recipe to Janus-Flow-1.3B extends it with any-to-any capabilities while preserving most of its original MMMU performance, at approximately $13\times$ less compute than the original Janus-Flow training. ``\xmark'' denotes capabilities not supported by the base model. $\downarrow$ Lower is better; $\uparrow$ Higher is better.}
\label{tab_supp:janus_flow_generalization}
\small
\setlength{\tabcolsep}{6pt}
\begin{tabular}{lccccc}
\toprule
Model & Scale & GH200 hrs & MMMU $\uparrow$ & DIODE $\downarrow$ & NYUv2 Surface Normal $\downarrow$ \\
\midrule
Janus-Flow (Base)               & 1.3B & 19.2k & 29.3 & \xmark & \xmark \\
Janus-Flow $+$ \modelname{}      & 1.3B & 1.5k  & 27.1 & 0.304  & 36.53 \\
4M                              & 2.8B & 16.9k & \xmark & 0.331  & 37.28 \\
\bottomrule
\end{tabular}
\end{table}

% \clearpage

\section{More Evaluations}
\label{sec:supp_evaluation}
\subsection{Depth Estimation}
\label{subsec:supp_depth}

\begin{table*}[t]
\centering
\caption{Zero-shot depth estimation comparison on NYUv2, ScanNet and DIODE benchmarks.}
\setlength{\tabcolsep}{8pt}
\renewcommand{\arraystretch}{1.2}

\resizebox{0.75\textwidth}{!}{
\begin{tabular}{l l | cc | cc | cc}
\toprule
\multirow{2}{*}{Group} & \multirow{2}{*}{Method} &
\multicolumn{2}{c|}{NYUv2} &
\multicolumn{2}{c|}{ScanNet} &
\multicolumn{2}{c}{DIODE} \\
& &
AbsRel$\downarrow$ & $\delta_1\uparrow$ &
AbsRel$\downarrow$ & $\delta_1\uparrow$ &
AbsRel$\downarrow$ & $\delta_1\uparrow$ \\
\midrule

% ---------------- Single-task ----------------
\multirow{5}{*}{Single-task}
& MiDaS & 0.111 & 0.885 & 0.121 & 0.846 & 0.332 & 0.715 \\
& Omnidata & 0.074 & 0.945 & 0.075 & 0.936 & 0.339 & 0.742 \\
& DPT-large & 0.098 & 0.903 & 0.082 & 0.934 & 0.182 & 0.758 \\
& DepthAnything & 0.043 & 0.980 & 0.043 & 0.981 & 0.261 & 0.759 \\
& DepthAnything v2 & 0.043 & 0.979 & 0.042 & 0.979 & 0.249 & 0.752 \\
\midrule

% ---------------- Encoder-Decoder ----------------
\multirow{2}{*}{Encoder--Decoder}
& Unified-IO & 0.059 & 0.970 & 0.063 & 0.965 & 0.369 & 0.708 \\
& 4M-XL & 0.068 & 0.951 & 0.065 & 0.955 & 0.331 & 0.734 \\
\midrule

% ---------------- Diffusion ----------------
\multirow{2}{*}{Diffusion}
& OneDiffusion & 0.087 & 0.924 & 0.094 & 0.906 & 0.399 & 0.661 \\
& DiCeption & 0.061 & 0.960 & 0.072 & 0.944 & 0.289 & 0.722 \\
\midrule

% ---------------- Decoder-only ----------------
\multirow{1}{*}{Decoder-only}
& \method (Ours) & 0.065 & 0.958 & 0.067 & 0.957 & 0.285 & 0.718 \\
\bottomrule
\end{tabular}
} % end resizebox
\label{tab_supp:depth}
\end{table*}

We present a detailed zero-shot depth estimation evaluation for \modelname{}. As shown in~\cref{tab_supp:depth}, we report results on three standard benchmarks, NYUv2, ScanNet, and DIODE, covering both indoor and outdoor scenes. \new{Despite being a unified decoder-only model trained jointly with many other modalities, \modelname{} delivers competitive depth performance across all datasets, on par with multitask encoder--decoder and diffusion-based baselines. Its performance is also comparable to specialized single-task depth estimators, suggesting that unified multimodal training does not compromise depth estimation quality.}

% \subsection{Surface Normal Estimation}

\subsection{Referring Object Grounding}
\label{subsec:supp_grounding}

\begin{table*}[t]
\centering
\caption{Zero-shot comparison on referring expression comprehension benchmarks (RefCOCO, RefCOCO+, RefCOCOg).}
\setlength{\tabcolsep}{6pt}
\renewcommand{\arraystretch}{1.25}

\resizebox{0.75\textwidth}{!}{
\begin{tabular}{l l | ccc | ccc | cc}
\toprule
\multirow{2}{*}{Group} & \multirow{2}{*}{Method} &
\multicolumn{3}{c|}{RefCOCO} &
\multicolumn{3}{c|}{RefCOCO+} &
\multicolumn{2}{c}{RefCOCOg} \\
& &
val & testA & testB &
val & testA & testB &
val & test \\
\midrule

% ---------------- Single-task models ----------------
\multirow{2}{*}{Single-task}
& GLIP-T     & 50.42 & 54.30 & 43.83 & 49.50 & 52.78 & 44.59 & 66.09 & 66.89 \\
& Grounding-DINO-T & 50.41 & 57.24 & 43.21 & 51.40 & 57.59 & 45.81 & 67.46 & 67.13 \\
\midrule

% ---------------- Decoder-only models ----------------
\multirow{2}{*}{Decoder-only}
& Kosmos-2        & 52.32 & 57.42 & 47.26 & 45.48 & 50.73 & 42.24 & 60.57 & 61.65 \\
& \method (Ours)    & 54.50  & 58.60  & 50.91  & 49.75  & 54.94  & 44.63  & 56.50  & 56.21 \\
\bottomrule

\end{tabular}
}
\label{tab_supp:grounding}
\end{table*}

We present zero-shot referring expression comprehension results in~\cref{tab_supp:grounding} on RefCOCO, RefCOCO+, and RefCOCOg. While prior specialist grounding models and existing decoder-only systems typically focus on a narrow set of tasks, \modelname{} achieves comparable or superior performance across all evaluation splits. Notably, Kosmos-2 is also a decoder-only model capable of both VQA and grounding, yet \modelname{} supports a far broader any-to-any generation setting while maintaining strong zero-shot grounding performance.

\begin{revised}
\subsection{Image--Text Capability Preservation}
\label{subsec:supp_image_text}

Extending a strong base model like BAGEL to many additional modalities risks degrading its original image--text capabilities. To mitigate this, \modelname{} mixes LLaVA-OneVision into \datasetname{} training. \Cref{tab_supp:image_text_preservation} reports a broader image--text evaluation beyond MMMU.

Training on \datasetname{} alone (without the LLaVA-OV mix) causes large drops across the board, \eg, POPE $87.23\to73.67$, VizWiz $59.41\to8.34$, and MME-S $2377\to1561$, confirming the importance of the data mixture. The full \modelname{} setup recovers most of these losses: VizWiz stays within $1.2$ points of BAGEL ($58.21$ vs $59.41$), POPE stays within $0.04$ of baseline, and MME scores remain within $3\%$ of BAGEL. Document and high-resolution benchmarks (DocVQA, ChartQA) show some residual drop, consistent with the LLaVA-OV mix not including BAGEL's original full VQA training mixture.
\end{revised}

\begin{table}[t]
\centering
\caption{\textbf{Preservation of image--text capabilities.} ``$+$ LLaVA-OV'' and ``$+$ \datasetname{} only'' are ablations that add the named data to BAGEL individually; \modelname{} combines BAGEL initialization with LLaVA-OneVision mixing and \datasetname{} training. Training on \datasetname{} alone causes large drops, while the full \modelname{} setup recovers most of BAGEL's original VQA and captioning capabilities. $\uparrow$ Higher is better.}
\label{tab_supp:image_text_preservation}
\small
\setlength{\tabcolsep}{4pt}
\begin{tabular}{lcccc}
\toprule
Benchmark & BAGEL & $+$ LLaVA-OV & $+$ \datasetname{} only & \textbf{\modelname{}} \\
\midrule
MMMU $\uparrow$    & 53.20   & 52.30   & 47.90   & 51.10 \\
MME-P $\uparrow$   & 1687.00 & 1663.56 & 1157.99 & 1637.69 \\
MME-S $\uparrow$   & 2377.17 & 2316.77 & 1560.85 & 2313.40 \\
POPE $\uparrow$    & 87.23   & 87.77   & 73.67   & 87.27 \\
VizWiz $\uparrow$  & 59.41   & 57.77   & 8.34    & 58.21 \\
DocVQA $\uparrow$  & 94.05   & 93.37   & 90.02   & 90.11 \\
ChartQA $\uparrow$ & 86.72   & 85.47   & 81.09   & 82.45 \\
\bottomrule
\end{tabular}
\end{table}

\begin{revised}
\subsection{Multi-Condition and Direct Any-to-Any Generation}
\label{subsec:supp_multi_cond}

Most standardized benchmarks are RGB- or text-centric. To probe \modelname{}'s behavior beyond these settings, we evaluate three families of mappings on DIODE depth and NYUv2 surface normal estimation: standard (RGB$\to$X), multi-condition (multiple modalities $\to$ X), and direct any-to-any between non-standard pairs (\cref{tab_supp:multi_cond_any2any}).

Multi-conditioning improves over the standard RGB$\to$Surface Normal baseline (RGB$+$Depth$\to$Surface Normal: $19.58$; Edge$+$Depth$\to$Surface Normal: $19.72$, vs $20.02$ standard). Notably, Edge$+$Depth$\to$Surface Normal approaches the RGB-based baseline without using RGB as input, suggesting that \modelname{} combines complementary signals across modalities rather than relying on RGB as a privileged input. Direct any-to-any transfers between sparser modality pairs are naturally weaker (Edge$\to$Depth $0.302$ vs RGB$\to$Depth $0.285$; Depth$\to$Surface Normal $29.51$ vs RGB$\to$Surface Normal $20.02$), but achievable within the same pretrained model and without task-specific pathways, which the underlying teacher models cannot do.
\end{revised}

\begin{table}[t]
\centering
\caption{\textbf{Multi-condition and any-to-any results beyond RGB/text-centric settings.} Standard, multi-condition, and direct any-to-any mappings on DIODE depth and NYUv2 surface normal estimation. Multi-conditioning improves over standard RGB$\to$Surface Normal; Edge$+$Depth$\to$Surface Normal is nearly as strong as the RGB-based mapping despite not using RGB as input. Direct any-to-any transfers between sparser modality pairs are naturally weaker but achievable within the same pretrained model. $\downarrow$ Lower is better.}
\label{tab_supp:multi_cond_any2any}
\small
\setlength{\tabcolsep}{6pt}
\begin{tabular}{lllc}
\toprule
Mapping & Source $\to$ Target & Dataset & \modelname{} $\downarrow$ \\
\midrule
\multirow{2}{*}{Standard}     & RGB $\to$ Depth          & DIODE & 0.285 \\
                              & RGB $\to$ Surface Normal         & NYUv2 & 20.02 \\
\midrule
\multirow{2}{*}{Multi-cond.}  & RGB $+$ Depth $\to$ Surface Normal  & NYUv2 & 19.58 \\
                              & Edge $+$ Depth $\to$ Surface Normal & NYUv2 & 19.72 \\
\midrule
\multirow{2}{*}{Any-to-any}   & Edge $\to$ Depth         & DIODE & 0.302 \\
                              & Depth $\to$ Surface Normal       & NYUv2 & 29.51 \\
\bottomrule
\end{tabular}
\end{table}

\begin{revised}
\subsection{Contamination Check}
\label{subsec:supp_contamination}

To ensure that \datasetname{}'s pseudo-label supervision does not produce evaluation contamination, we checked the overlap between teacher training data and our evaluation benchmarks. \textbf{Depth:} DepthAnything V2~\cite{yang2024depth} and Marigold~\cite{ke2025marigold} are not trained on DIODE or NYUv2. \textbf{Grounding:} our GLaMM-based pipeline~\cite{rasheed2024glamm} uses a subset of SA-1B and remains zero-shot on COCO, which is not used during \datasetname{} construction. \textbf{Features:} DINOv2~\cite{oquab2023dinov2} is not trained on the ImageNet splits used for our retrieval evaluation. \textbf{Source images:} \datasetname{} is built on BLIP-3o sources (SA-1B, JourneyDB, CC12M), none of which directly overlap with our evaluation benchmarks. Together, these checks indicate that \modelname{}'s reported numbers reflect cross-modal modeling rather than memorization of teacher behavior.
\end{revised}

\begin{revised}
\subsection{GenEval Per-Category Breakdown and Self-Verification}
\label{subsec:supp_geneval}

To analyze where \modelname{} gains and loses relative to BAGEL on text-to-image generation, we report a per-category GenEval breakdown in \cref{tab_supp:geneval_breakdown}. \modelname{} matches or slightly improves on simpler categories (single-object, colors) but loses on harder compositional ones (counting, position, color$+$position). A simple Best-of-4 test-time search using \modelname{}'s own grounding and VQA capabilities to score candidate generations recovers most of this gap, lifting overall GenEval from $0.81$ to $0.84$. This is a direct consequence of the unified design: \modelname{} can act as its own verifier within the same model, without external scoring components.

\paragraph{Connection to test-time scaling for image generation.} Best-of-$N$ verification is part of a broader emerging direction of test-time scaling for image generation, where additional inference compute substitutes for additional training. Prior work explores this through repeated sampling with verifiers~\cite{brown2024large,snell2024scaling}, inference-time search tailored to diffusion models~\cite{ma2025its,singhal2025general,zhang2025classical}, and tokenizer-aware ordered-token search for autoregressive image generators~\cite{soto,chen2025tts}. Our self-verification result shows that a unified any-to-any model is naturally suited to this paradigm: the same decoder that generates an image also produces the auxiliary modalities (grounding, VQA) used to score it, removing the need for any external verifier.
\end{revised}

\begin{table}[t]
\centering
\caption{\textbf{GenEval breakdown by category and effect of self-verification.} Compared with BAGEL, \modelname{} matches or slightly improves on simple categories (single-object, colors) but loses on harder compositional ones (counting, position, color$+$position). A simple Best-of-4 self-verification (re-ranking with \modelname{}'s own grounding and VQA) recovers most of the gap. $\uparrow$ Higher is better.}
\label{tab_supp:geneval_breakdown}
\small
\setlength{\tabcolsep}{4pt}
\begin{tabular}{lccccccc}
\toprule
Model & Overall & Single Obj. & Two Obj. & Count & Colors & Position & Color$+$Pos. \\
\midrule
BAGEL                                & 0.86 & 0.98 & 0.94 & 0.74 & 0.94 & 0.84 & 0.74 \\
\modelname{}                          & 0.81 & 0.99 & 0.90 & 0.68 & 0.92 & 0.77 & 0.62 \\
\modelname{} (Best-of-4, self-verify) & 0.84 & 1.00 & 0.90 & 0.75 & 0.94 & 0.80 & 0.65 \\
\bottomrule
\end{tabular}
\end{table}

\begin{revised}
\subsection{Inference Efficiency: Independent vs Chained}
\label{subsec:supp_inference}

\modelname{}'s 2D Expert uses parallelized flow matching rather than token-by-token autoregression, so chained generation does not introduce a prohibitive latency cost. \Cref{tab_supp:inference_efficiency} benchmarks RGB$\to$Surface Normal and RGB$\to$Canny$\to$Surface Normal at $512\times512$ resolution on a single GH200 GPU over the 654-image NYUv2 split. Two observations: (i) independent generation is already accurate at $5$--$10$ steps; (ii) chained generation reaches accuracy comparable to many-step independent generation at substantially lower latency, \eg, 10-step chained ($4.28$ s/img, $20.08$ surface normal error) matches 50-step independent ($9.07$ s/img, $20.02$ surface normal error) at less than half the latency. This trade-off makes chained generation practical even when intermediate modalities are added.
\end{revised}

\begin{table}[t]
\centering
\caption{\textbf{Inference efficiency: independent vs chained generation.} Latency and NYUv2 surface normal estimation accuracy at $512\times512$ resolution on a single GH200 GPU over 654 NYUv2 images. Chained generation does not introduce a prohibitive autoregressive bottleneck: 10-step chained (4.28 s/img) reaches the accuracy of 50-step independent (9.07 s/img) at less than half the latency. $\downarrow$ Lower is better.}
\label{tab_supp:inference_efficiency}
\small
\setlength{\tabcolsep}{6pt}
\begin{tabular}{llcc}
\toprule
Task & Steps & Sec/img & NYUv2 Surface Normal $\downarrow$ \\
\midrule
\multirow{4}{*}{RGB $\to$ Surface Normal}                 & 2  & 0.59  & 34.06 \\
                                                   & 5  & 1.14  & 21.91 \\
                                                   & 10 & 2.05  & 20.30 \\
                                                   & 50 & 9.07  & 20.02 \\
\midrule
\multirow{4}{*}{RGB $\to$ Canny $\to$ Surface Normal}      & 2  & 1.38  & 36.82 \\
                                                   & 5  & 2.57  & 22.21 \\
                                                   & 10 & 4.28  & 20.08 \\
                                                   & 50 & 20.05 & 19.87 \\
\bottomrule
\end{tabular}
\end{table}

\clearpage

\section{More Visualizations}
\label{sec:supp_visualization}

\subsection{Independent Any-to-Any Generation}
\label{subsec:supp_independent}

\begin{figure*}[t!]
  \centering
  \includegraphics[width=\textwidth]{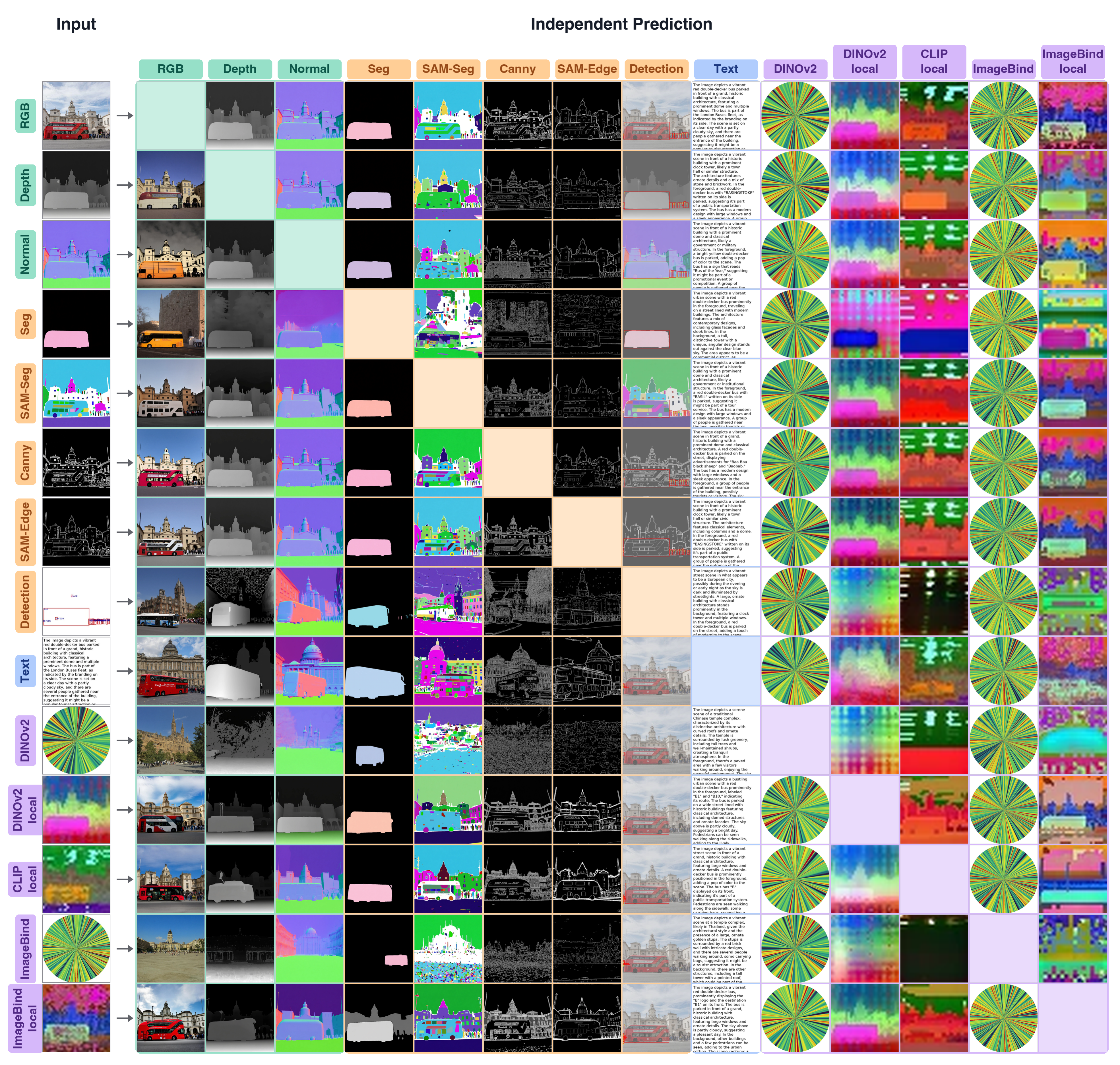}
  \vspace{-0.1in}
  \caption{\textbf{Any-to-Any Independent Generation.} Rows are the input modality (leftmost column) and columns are the target; each off-diagonal cell shows \modelname{} generating the column's modality directly from the row's modality within a single unified architecture (the diagonal, input $=$ target, is left blank). Each target is predicted independently from the input, so the outputs along a row are free to vary and need not agree on a single scene. The chained counterpart used in the main paper, which conditions each output on the previous ones to enforce consistency, is shown in \cref{fig:chain_gen}. Interactive visualizations are available at \href{https://modus-multimodal.epfl.ch/\#any-to-any}{modus-multimodal.epfl.ch}.
  }
  \label{fig:any2any_vis}
  \vspace{-0.2in}
\end{figure*}

As shown in \cref{fig:any2any_vis}, \modelname{} can take an arbitrary input modality and generate all other target modalities within the same architecture, demonstrating any-to-any translation across diverse representations. Here each target is generated independently from the input, so the outputs are not constrained to be mutually consistent; the chained variant used in the main paper~(\cref{fig:chain_gen}) instead conditions each output on the previously generated ones to enforce cross-modal consistency.

\subsection{Chained Generation}
\label{subsec:supp_chained_gen}

\begin{figure}[h!]
  \centering
  \includegraphics[width=0.95\columnwidth]{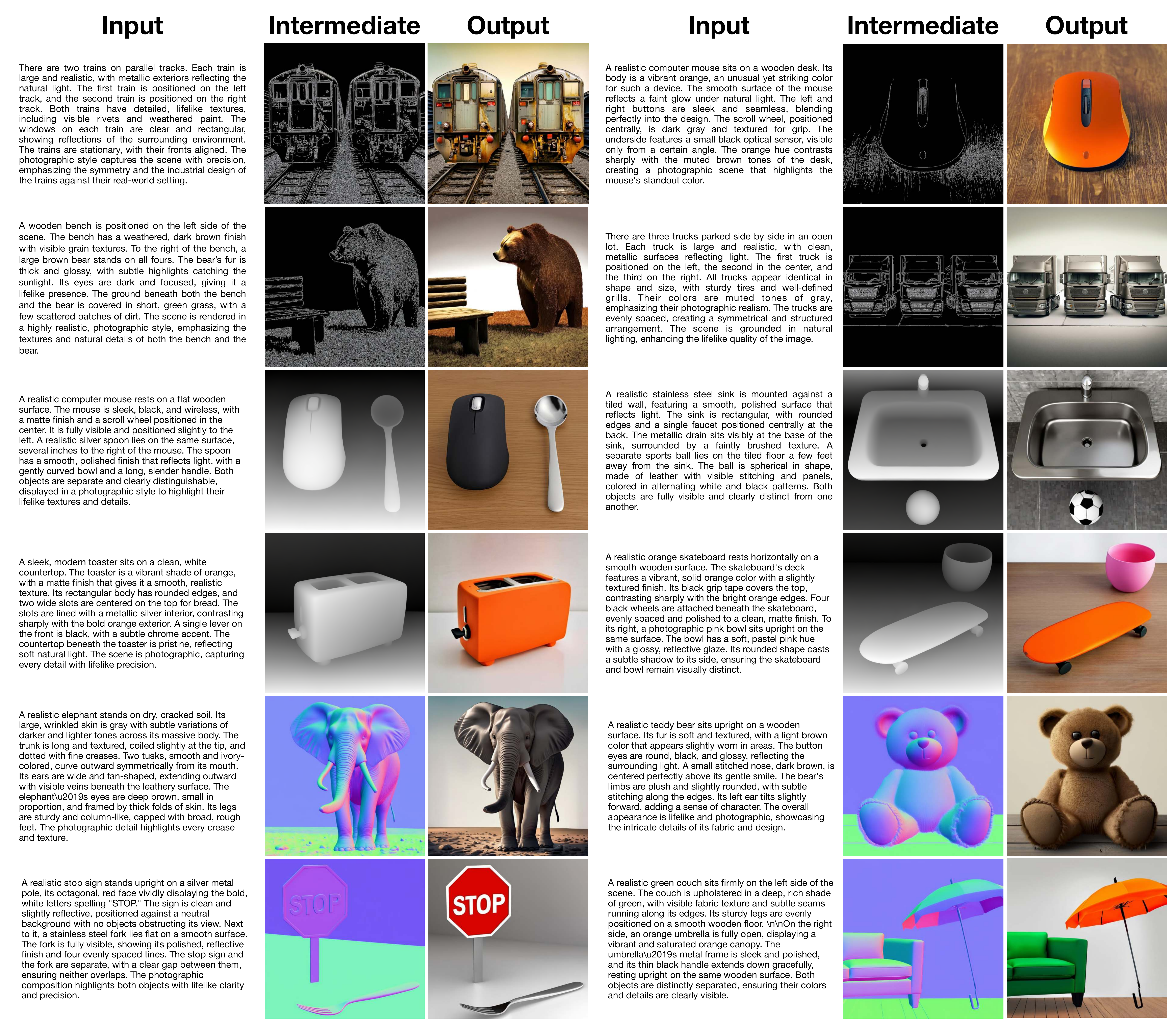}
\caption{
\textbf{Chained text-to-image generation.}
Text prompts are transformed into intermediate 2D modalities, including Canny edges, depth maps, and surface normals, before producing the final RGB image. The examples demonstrate high visual quality and strong cross-modality consistency throughout the chained generation process.
}
  \label{fig_supp:chain_gen}
\end{figure}

\begin{figure}[h!]
  \centering
  \includegraphics[width=0.95\columnwidth]{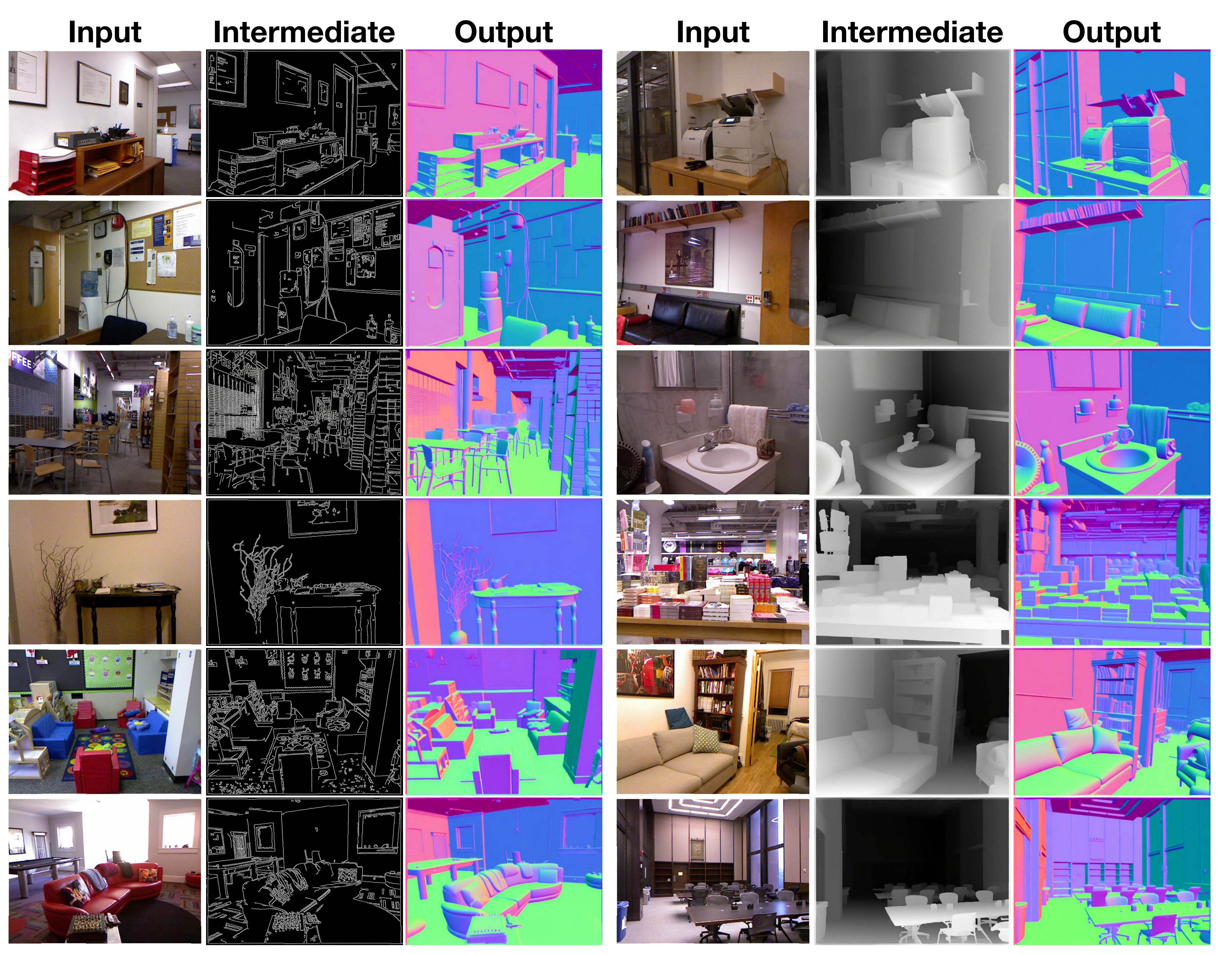}
\caption{
\textbf{Chained image-to-surface normal prediction.}
The input image is first transformed into intermediate modalities, such as depth or Canny edges, and the resulting representations are then used to produce the final surface normal map. The examples illustrate consistent and coherent predictions across the chained modalities.
}
  \label{fig_supp:chain_normal}
\end{figure}

In Section 4.3 of the main paper, we discussed chained generation, where \modelname{} produces intermediate modalities before generating the final target output. In~\cref{fig_supp:chain_gen} and~\cref{fig_supp:chain_normal}, we include additional visualizations to further illustrate this capability. Across tasks such as text to image and image to surface normal, we observe that intermediate predictions, including Canny edges, depth maps, and surface normals, remain coherent and consistent with the source input. The final outputs follow these intermediate representations closely, showing that \modelname{} can maintain structural and semantic consistency throughout the chained generation process.

\clearpage 

\subsection{Self-Verification}
\label{subsec:self-verify}

\begin{figure}[h!]
  \centering
  \includegraphics[width=0.5\columnwidth]{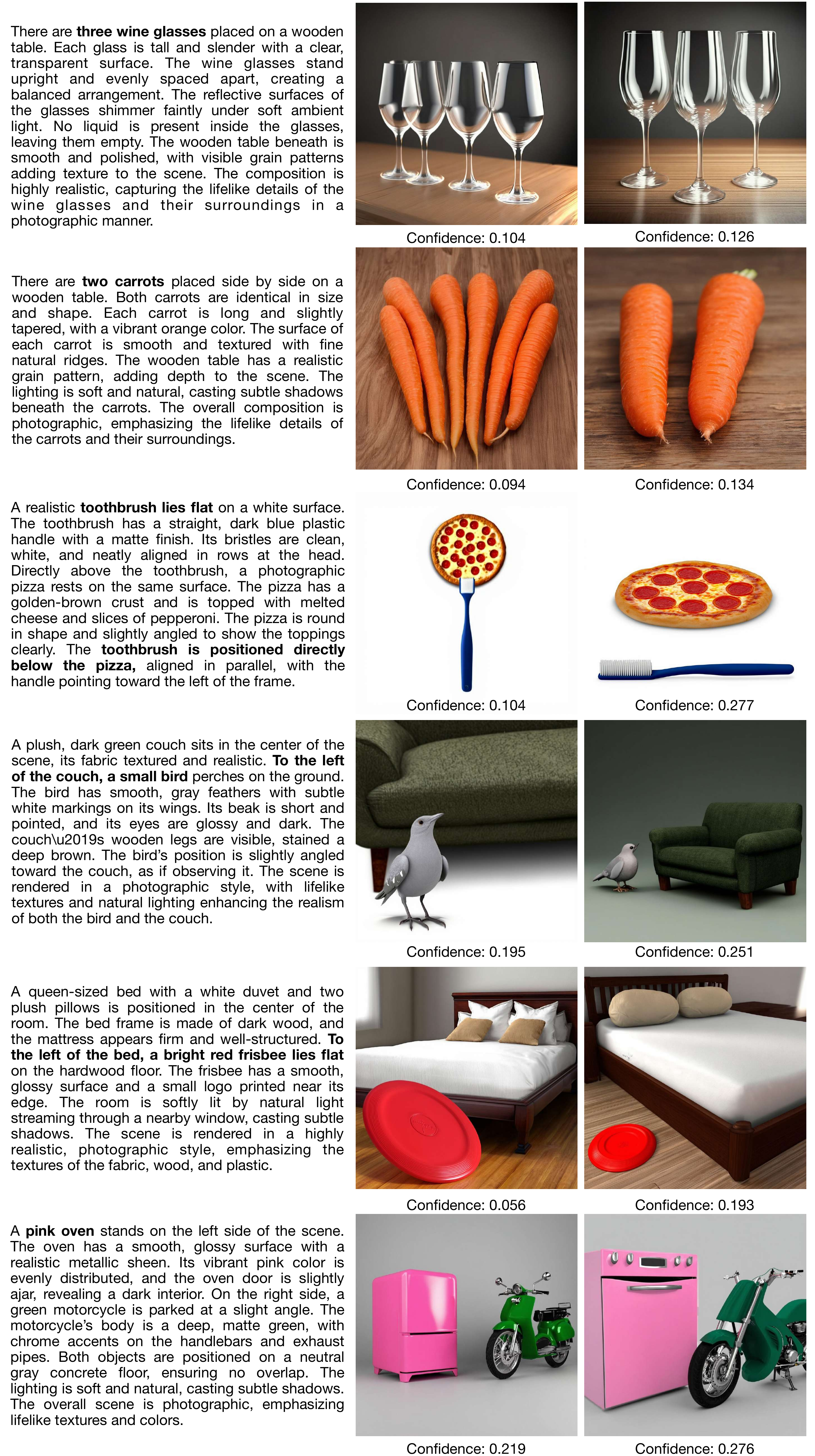}
  \vspace{-0.1in}
\caption{
\textbf{Text-to-image generation with self-verification.}
We apply the grounding capability of \modelname{} to evaluate the quality of its own text-to-image outputs and select the sample with the highest verification score. As shown, this simple test-time search, using a task already supported by \modelname{}, leads to improved image quality and better alignment with the input prompt.
}
  \label{fig_supp:verifier}
\end{figure}

In Sec. 4.4 of the main paper, we discussed how the referring object grounding modality in \modelname{} provides verification for text-to-image generation. Here, we present additional visualizations in~\cref{fig_supp:verifier}. For each text prompt, we generate several image candidates and simply read the grounding logits produced by \modelname{} when asked to localize the referred objects. We observe that these confidence values often correlate with whether the generated image includes the requested objects and with cues such as approximate count or location. Based on this observation, we choose the sample with the highest grounding confidence as the representative output. The visualizations illustrate that grounding confidences can provide a simple signal for examining object presence and prompt adherence in text-to-image generations, without introducing additional training or external scoring models.

\subsection{Visual Representation Composition}
\label{subsec:supp_visual_representation}

\begin{figure}[h!]
  \centering
  \includegraphics[width=0.95\columnwidth]{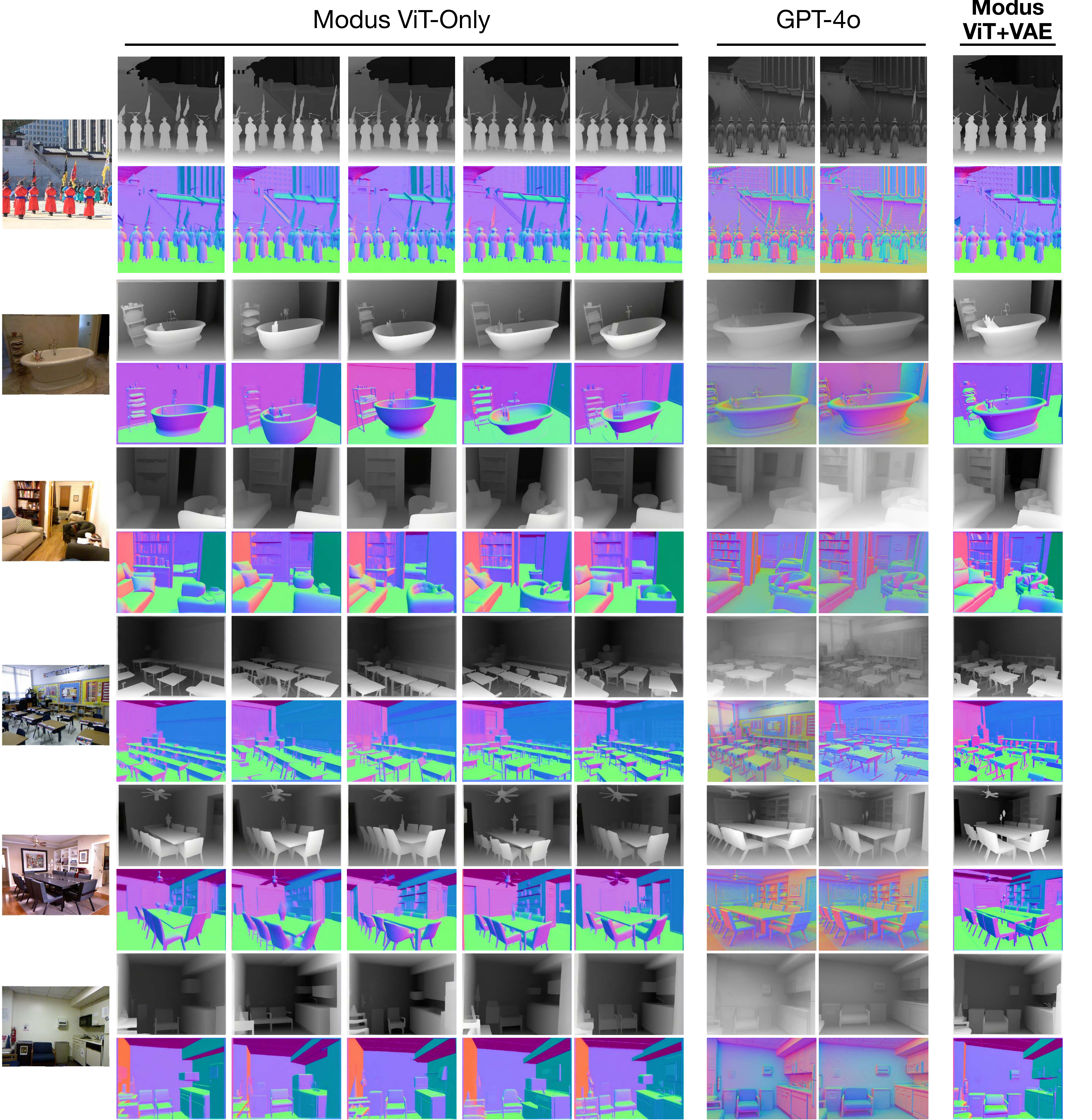}
\caption{
\textbf{Illustration of hallucination with 2D modality feature representations.}
Each 2D modality in \modelname{} is represented using a ViT-based semantic feature and a VAE-based reconstruction feature. Conditioning the generation on the semantic feature can produce plausible but structurally incorrect depth or surface normal maps, showing changes in object shape, number, or fine details. GPT-4o exhibits similar behavior when asked to generate depth maps, including mild modality confusion. The final column shows accurate results from \modelname{} when both semantic and reconstruction features are used as conditions.
}
  \label{fig_supp:vit_only}
\end{figure}

In Sec. 4.5 of the main paper, we reported an intriguing form of visual hallucination that emerges when \modelname{} generates 2D modalities using only high-level semantic features. In \modelname{}, each 2D modality is represented using two complementary encodings: (1) a ViT-derived semantic feature for global understanding, and (2) a VAE-based reconstruction feature that preserves local geometry and fine details. To better illustrate the impact of these two branches, we provide additional qualitative results in~\cref{fig_supp:vit_only}.

When we condition the generation on ViT features, the model is still able to produce plausible depth or surface normal maps that are semantically consistent with the input image. However, these outputs often deviate from the exact scene structure. Across multiple samples per image, five for depth and five for surface normals, we observe systematic distortions: object shapes may change, the number of instances may vary, and fine-grained geometric details may be altered or lost. These variations indicate that high-level semantic features alone do not sufficiently constrain the spatial layout, allowing the model to generate semantically aligned but structurally inconsistent predictions.

Interestingly, this phenomenon echoes observations reported in recent evaluations of GPT-4o's image generation capabilities~\cite{ramachandran2025doesgpt4ounderstandvision}. We collected GPT-4o outputs using prompts such as “Please generate the gray-scale depth map of the input image and keep the original resolution.” Similar hallucinations occur: missing object parts, changes in inferred camera viewpoint, and in some cases mild modality confusion that may stem from limited training coverage on dense geometric modalities. These parallels are consistent with GPT-4o relying on a similar higher-level, semantic conditioning for structure-sensitive outputs, though we cannot inspect its internals. It also suggests that high-level–only conditioning, even in large proprietary systems, may drive a form of semantic understanding rather than precise geometric prediction.

In the final column of~\cref{fig_supp:vit_only}, we show the corresponding results from \modelname{} when conditioned on both ViT semantic features and VAE low-level features. This combination reinstates geometric fidelity, producing depth and surface normal maps that accurately preserve scene structure, object boundaries, and local surface details. Overall, these comparisons highlight the importance of coupling high-level semantics with low-level geometric cues in unified multimodal generation, and they provide insight into why models relying solely on high-level conditioning may exhibit consistent visual hallucinations.

\clearpage

\section{Implementation Details}
\label{sec:supp_implementation}

\subsection{Dataset}
We train \modelname{} on \datasetname{}, a large-scale multimodal corpus of 29M samples constructed by extending the BLIP-3o~\cite{chen2025blip3} image--caption corpus with aligned annotations across diverse modalities. To obtain supervision at this scale, especially for modalities where human annotations are scarce, we apply high-quality pseudo-labelers with an emphasis on both accuracy and efficiency. Specifically, we use Grounded-SAM~\cite{ren2024grounded} and SAM~\cite{kirillov2023segment} for instance segmentation, DepthAnything V2~\cite{yang2024depth} for depth estimation, and Marigold~\cite{ke2025marigold} for surface normal prediction.

We also generate edge maps using the classical Canny operator and SAM~\cite{kirillov2023segment}, and extract global and spatially-local feature maps using DINOv2~\cite{oquab2023dinov2}, CLIP~\cite{clip}, and ImageBind~\cite{girdhar2023imagebind}. For localization supervision, we use ground truth bounding boxes from the GLaMM~\cite{rasheed2024glamm} dataset for grounding and ViTDet~\cite{li2022vitdet} with an EVA-02 backbone~\cite{fang2024eva02} for detection, both overlapping with the BLIP-3o data sources and providing reliable annotations. In addition, we incorporate the LLaVA-OneVision~\cite{Li2024LLaVAOneVisionEV} dataset to improve VQA training for the text modality.

\Cref{tab_supp:modality_overview} summarizes the full set of modalities supported by \modelname{}, organized by family and by generation mechanism. The set is representative rather than fixed: each family can host multiple instantiations (\eg, classical and SAM-based edges, or global and spatially-local feature maps), and additional modalities can be incorporated without architectural changes. All modalities follow the same unified tokenization and training recipe.

\begin{table*}[t]
\centering
\caption{\textbf{Overview of modalities supported by \modelname{}.} Modalities are organized by family and by generation mechanism (1D autoregressive next-token prediction vs.\ 2D flow matching). The set is representative rather than fixed: each family can host multiple instantiations, and new modalities can be added without architectural changes.}
\setlength{\tabcolsep}{6pt}
\renewcommand{\arraystretch}{1.25}
\resizebox{0.92\textwidth}{!}{
\begin{tabular}{l l c l l}
\toprule
Modality & Family & Dim & Representation & Supervision / source \\
\midrule
\multicolumn{5}{l}{\textit{1D sequential modalities (autoregressive next-token prediction)}} \\
\midrule
Text                & Language          & 1D & text tokens             & BLIP-3o~\cite{chen2025blip3}, LLaVA-OneVision~\cite{Li2024LLaVAOneVisionEV} \\
Visual Grounding    & Localization      & 1D & box tokens              & GLaMM~\cite{rasheed2024glamm} \\
Detection           & Localization      & 1D & box + label tokens      & ViTDet~\cite{li2022vitdet} w/ EVA-02~\cite{fang2024eva02} \\
DINOv2 (global)     & Learned features  & 1D & feature tokens          & DINOv2~\cite{oquab2023dinov2} \\
ImageBind (global)  & Learned features  & 1D & feature tokens          & ImageBind~\cite{girdhar2023imagebind} \\
CLIP (local)        & Learned features  & 1D & spatial feature tokens          & CLIP~\cite{clip} \\
DINOv2 (local)      & Learned features  & 1D & spatial feature tokens  & DINOv2~\cite{oquab2023dinov2} \\
ImageBind (local)   & Learned features  & 1D & spatial feature tokens  & ImageBind~\cite{girdhar2023imagebind} \\
\midrule
\multicolumn{5}{l}{\textit{2D spatial modalities (flow matching in latent space)}} \\
\midrule
RGB                 & Appearance        & 2D & VAE + ViT               & BLIP-3o~\cite{chen2025blip3} \\
Depth               & Geometry          & 2D & VAE + ViT               & DepthAnything V2~\cite{yang2024depth} \\
Surface Normal      & Geometry          & 2D & VAE + ViT               & Marigold~\cite{ke2025marigold} \\
Segmentation        & Semantic          & 2D & VAE + ViT               & Grounded-SAM~\cite{ren2024grounded} \\
SAM-Seg             & Semantic          & 2D & VAE + ViT               & SAM~\cite{kirillov2023segment} \\
Canny               & Structure         & 2D & VAE + ViT               & classical Canny operator \\
SAM-Edge            & Structure         & 2D & VAE + ViT               & SAM~\cite{kirillov2023segment} \\
\bottomrule
\end{tabular}
}
\label{tab_supp:modality_overview}
\end{table*}

\subsection{Training}
During training, we sample a sequence of modalities and feed them to the model. Each sequence contains several conditioning modalities and one target modality. The attention pattern between different modalities is kept causal. Within each modality, 1D modalities use causal attention, while 2D modalities use bidirectional attention for both ViT tokens and VAE tokens.
We also apply sequence packing to maintain efficient and stable sequence lengths during training. Depending on the type of the target modality, we compute its loss using cross entropy for 1D tokens and mean squared error for 2D features.
Detailed training configuration for different stages is shown in~\cref{supp_table:config}.

\begin{table*}[t]
\centering
\caption{\textbf{Training settings.} Training Configuration for \method used in three stages.}
\setlength{\tabcolsep}{6pt}
\renewcommand{\arraystretch}{1.2}

\begin{tabular}{l|
    >{\centering\arraybackslash}p{3cm}
    >{\centering\arraybackslash}p{3cm}
    >{\centering\arraybackslash}p{3cm}}
\toprule
Configuration & Stage 1 & Stage 2 & Stage 3 \\
\midrule
Training length & 30B & 20B & 15B \\
Warmup length &  & 1B &  \\
Optimizer &  & AdamW &  \\
Opt.\ momentum &  & $\beta_1,\beta_2 = 0.9, 0.95$ & \\
Base learning rate &  & 2e-5 &  \\
Sequence Length & 14336 & 14336 & 16384 \\
Weight decay &  & 0 &  \\
Gradient clipping &  & 1.0 &  \\
Learning rate schedule &  & Constant &  \\
\midrule
Num.\ of condition modalities & 1 & 1 & 3 \\
Modalities & RGB, Text, Grounding, DINOv2 & All & All \\
\midrule
ViT max resolution &  & (224, 518) &  \\
VAE resolution &  & (512, 1024) & \\
Augmentation &  & None &  \\
Data type &  & bfloat16 &  \\
\bottomrule
\end{tabular}

\label{supp_table:config}
\end{table*}

\section{Limitation Discussion}
\label{sec:supp_limitation}

While \modelname{} is designed as a unified decoder-only model capable of any-to-any multimodal generation, several aspects remain open for future improvement.

First, \modelname{} focuses on transformations across different modalities, but it does not yet support \emph{iterative editing within the same modality} (\eg, depth~$\rightarrow$ refined depth, segmentation~$\rightarrow$ edited segmentation). This capability requires curated datasets that contain sequential edits or multi-turn modality refinements, which are scarce in existing multimodal corpora. We believe that a unified any-to-any model is a useful step toward enabling the creation of such synthetic multimodal editing datasets, which in turn can feed back into future training to improve the model’s ability to handle iterative within-modality edits.

Second, \modelname{} is trained purely in a pre-training setting and therefore does not explicitly target reasoning tasks. While the model already demonstrates generalization across modalities, supporting tasks that require complex reasoning would benefit from incorporating dedicated post-training objectives and datasets, such as multimodal chain-of-thought (\eg, DeepEyes~\cite{zheng2025deepeyes}). Extending \modelname{} with a lightweight post-training stage is a natural next step and would broaden its utility without altering the unified architecture.

Overall, these limitations highlight opportunities for improvement in dataset coverage and task formulation. Addressing them will require further investigation, but the results suggest that there remains strong potential to support an even broader range of tasks within a single unified decoder framework.

\clearpage

\section{\datasetname{} Examples}
\label{sec:pseudo_label_vis}

\begin{figure*}[h]
    \centering
    \includegraphics[width=\textwidth]{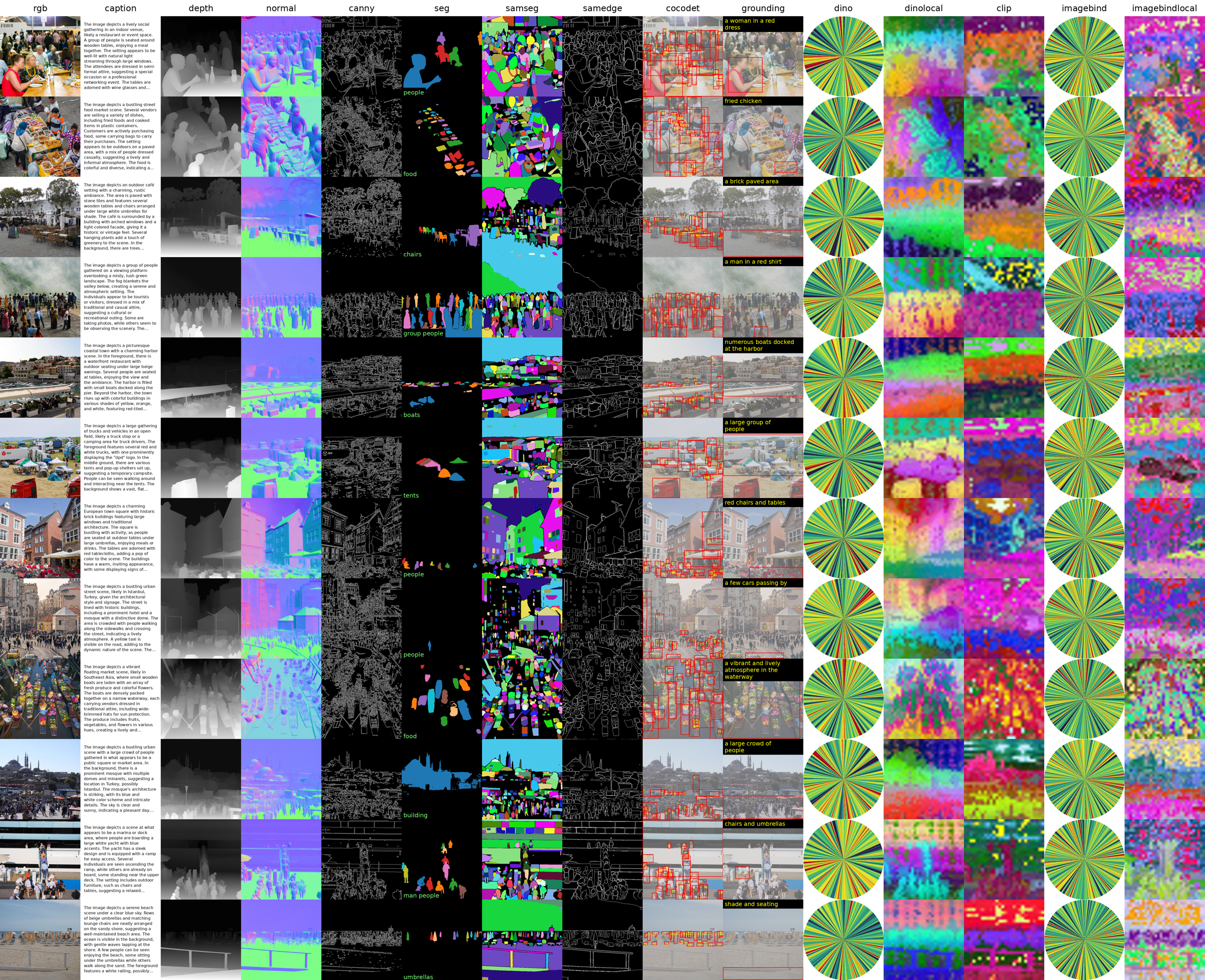}
    \caption{
    \textbf{\datasetname{} examples.} Each row shows one image together with its aligned annotations across the supported modalities: RGB, caption, depth, surface normal, canny and SAM-based edges, segmentation and SAM-based masks, detection, grounding, and DINOv2, CLIP, and ImageBind features (in both global and local forms).
    }
    \label{fig:appendix_pseudo_label_vis_1}
\end{figure*}

We present example visualizations of pseudo labels from \datasetname{} in~\cref{fig:appendix_pseudo_label_vis_1}. We release the full \datasetname{} together with two \modelname{} checkpoints, \texttt{Modus-15modality-14B-A7B} and \texttt{Modus-15modality-77B-A13B}.

\end{document}